%% file: main.tex
\documentclass[sigconf]{acmart}

\usepackage{enumerate}
\usepackage{tcolorbox}
\usepackage{subcaption}
\usepackage{multirow}
\usepackage{makecell}
\usepackage{tabularx}
\usepackage{algorithm}
\usepackage{algorithmic}
\usepackage{stfloats}
\tcbuselibrary{listings,breakable,skins}
\definecolor{aclback}{RGB}{245,245,250}
\definecolor{aclbluer}{RGB}{0,102,204}
\newcommand{\method}{Net-Ev$^2$}
\newcommand{\task}{Net-Ev$^2$}

\AtBeginDocument{%
  }

\copyrightyear{2026}
\acmYear{2026}
\setcopyright{cc}
\setcctype{by}
\acmConference[KDD 2026] {Proceedings of the 32nd ACM SIGKDD Conference on Knowledge Discovery and Data Mining V.2}{August 9--13, 2026}{Jeju Island, Republic of Korea.}
\acmBooktitle{Proceedings of the 32nd ACM SIGKDD Conference on Knowledge Discovery and Data Mining V.2 (KDD 2026), August 9--13, 2026, Jeju Island, Republic of Korea}
\acmISBN{979-8-4007-2259-2/2026/08}
\acmDOI{10.1145/3770855.3817972}

\begin{document}
% Jilin hu 
\title{Net-Ev$^2$: A Generative Simulator for Network Event Evolution}

% Net-Ev^2: a generative simulator for network event evolution
% Net-Ev^2: externally caused network event evolution - externally can be naturally defined as a set of conditions

% Net-Ev^2: network event evolution simulation - evolution simulation implies evolution driven by certain conditions

% Net-Ev^2 NetEvo

% Event-condition, Topological Generator, Spatial-Temporal Data


\author{Guangyu Wang}
\affiliation{%
  \institution{NYU Shanghai}
  \city{Shanghai}
  \country{China}
}
\email{gw2556@nyu.edu}

\author{Zhaonan Wang}
\authornote{Corresponding author, also affiliated with NYU Shanghai Center for Data Science and Shanghai Key Laboratory of Urban Design and Urban Science.}
\affiliation{%
  \institution{NYU Shanghai}
  \city{Shanghai}
  \country{China}
}
\email{zhaonan.wang@nyu.edu}
% emails go last

\renewcommand{\shortauthors}{Guangyu Wang and Zhaonan Wang}
% https://www.scomminc.com/pp/acmsig/kdd-papers.htm#L
\begin{abstract}
% Network dynamic evolution under event conditions generation aims to synthesize realistic traffic patterns on road networks under specific event scenarios, enabling simulation-based decision support and data augmentation for urban traffic systems. 
% % This is a nontrivial task that introduces several new challenges.

Reducing real-world trial and error has long been a central goal of decision making, and generative simulators advance this goal by modeling the evolution of future states.
An even more challenging yet meaningful task is simulating how disturbance events (e.g., accidents) propagate their impacts across real-world networks. 
The existing approaches fall short of modeling both structured attributes and unstructured semantics of events, and capturing topological structures in simulating network event evolution.
Therefore, we are motivated to propose \task{} (\underline{\textbf{Net}}work \underline{\textbf{Ev}}ent \underline{\textbf{Ev}}olution), a novel generative simulator that jointly leverages event cues while preserving network topology in simulations. 
Specifically, the framework consists of two stages, namely structure-guided masked pre-training and topology-aware diffusion process, which is achieved by U-Net-like graph downsampling and upsampling during denoising.
%We first train an encoder with a structure-guided masking objective to learn a latent representation for subsequent generation.
%In latent space, the generator evolves network states via graph pooling and unpooling operations with unstructured condition modulation.
At inference time, \method can generate simulations using natural-language event input only, with greater flexibility for practical usage.
% the model is generated using only unstructured event descriptions for greater flexibility.
Furthermore, we introduce \task{}-6.5M, a multimodal benchmark of aligned event and network traffic data across four large-scale road networks, as well as a new topology-aware metric, namely JL-MMD, to evaluate topological fidelity in generated network dynamics. % motivated by the Johnson–Lindenstrauss lemma
Extensive experiments demonstrate the state-of-the-art performance and strong generalization ability of \method.
Code is made available at \url{https://github.com/Guangyu4/Net-Ev-2}.

\end{abstract}
% Net-Ev^2-6.5M





\maketitle

\newcommand\kddavailabilityurl{https://doi.org/10.5281/zenodo.20506368}
\ifdefempty{\kddavailabilityurl}{}{
\begingroup\small\noindent\raggedright\textbf{Resource Availability:}\\
% please change the following context to include multiple artifacts if necessary, including data, models, code, etc.
The source code of this paper has been made publicly available at \url{\kddavailabilityurl}.
\endgroup
}
\input{content/1_Introduction}
\input{content/3_Preliminaries}
\input{content/4_Methodology}
\input{content/5_Experiments}
\input{content/6_Related_Work}
\input{content/7_Conclusion}

\begin{acks}
This work was supported by Shanghai Pujiang Program (STCSM No. 24PJA094). The authors also gratefully acknowledge the high-performance computing (HPC) support provided by NYU Shanghai.
\end{acks}
\balance
% \newpage
\bibliographystyle{ACM-Reference-Format}
% \balance
\bibliography{sample-base}

\input{content/9_arxiv}  % appendix goes after references

\end{document}

%% file: content/1_Introduction.tex
\section{Introduction}
% Diffusion topological unstructured and structured

% Generative models, particularly diffusion models, have emerged as a powerful paradigm for data synthesis, demonstrating remarkable capabilities such as text-to-image~\cite{li2019controllable}, text-to-audio~\cite{huang2023make,liu2023audioldm}, and text-to-video generation~\cite{singer2022make}. 

Generative simulators model state evolution over time~\cite{pmlr-v119-sanchez-gonzalez20a}, enabling closed-loop evaluation~\cite{tan2024promptable,ajay2023is} and scalable data generation~\cite{NEURIPS2023_d95cb79a} to reduce real-world trial and error in decision-making.
An even more challenging yet meaningful objective is to simulate real-world responses under event conditions~\cite{pmlr-v97-oberst19a,10.5555/3495724.3497532}, such as network event evolution~\cite{zhang2020curb}. 
Since events can be hard to predict~\cite{ghil2011extreme,ding2019modeling}, it is important to understand how they can affect a system over time.

Toward this goal, existing generative models can be broadly grouped into two design paradigms. (\romannumeral1) The first paradigm converts the network traffic into image-like representations to leverage advances in visual diffusion models. 
Representative methods use mappings such as line graphs~\cite{li2023time}, field transforms~\cite{naiman2024utilizing}, or directly render data as RGB images by combining multiple variables~\cite{zhang2025chattraffic}. 
Although they achieve promising results in some settings, these methods often struggle to capture the full complexity of network evolution. 
(\romannumeral2) The second paradigm adapts diffusion architectures such as flow matching~\cite{lipman2022flow} and Denoising Diffusion Probabilistic Models (DDPM)~\cite{ho2020denoising} to better match real-world properties, including variable-length sequences~\cite{ge2025t2s}, multivariate dependencies~\cite{gu2025verbalts}, and long-term trends~\cite{yuan2024diffusionts}.

\begin{figure}[ht]
    \centering
    \includegraphics[width=0.98\linewidth]{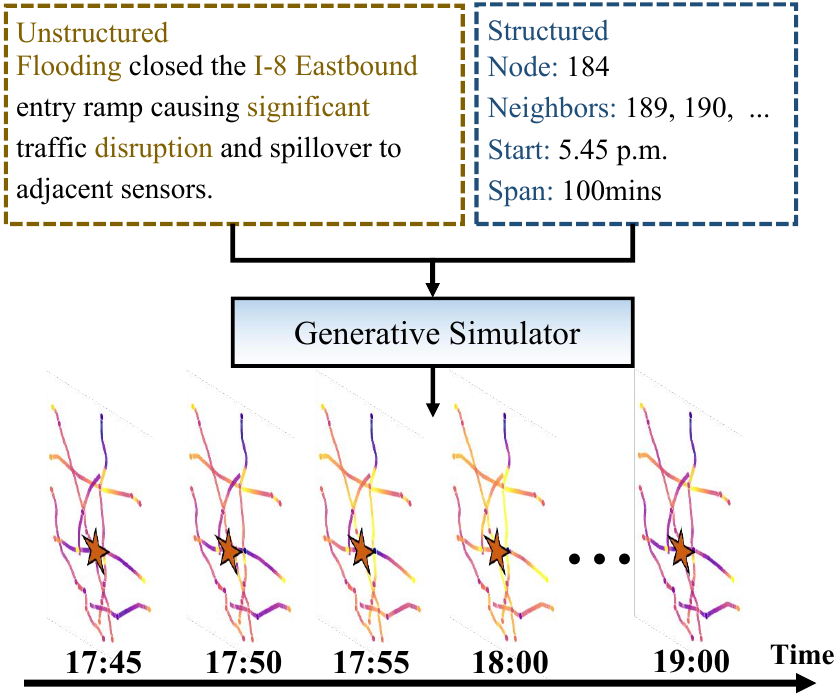}
    \caption{Generative simulation for network event evolution}
    \Description{An event description with unstructured text and structured node and time attributes is fed into a generative simulator, which outputs a sequence of network traffic states over time. The illustrated traffic maps show event effects evolving from 17:45 to 19:00 around the affected location.}
    % takes diverse event attributes as conditions to generate future network evolution. Colors indicate traffic values. At 19:00, traffic returns to a level close to 17:45, suggesting that the earlier change is caused by an event.
    \label{fig:teaser}
\end{figure}

However, network event evolution poses new challenges for existing methods, as illustrated in Figure~\ref{fig:teaser}. 
(1) Events have multi-structured attributes across both text and numbers, and these two modalities provide complementary cues. For example, attributes such as location and severity can be naturally described in unstructured text (e.g., I-8 Eastbound'' and significant''), while numerical attributes such as node indices and time spans are structured and not easily expressed in text.
Directly merging these features into conditioning input can be problematic, since language encoders like BERT~\cite{devlin-etal-2019-bert} are designed for semantic rather than numerical information.
As a result, structured attributes are often poorly represented and bring only marginal performance gains~\cite{parihar2025kontinuous}.
Moreover, structured data is often harder to obtain than unstructured data in real-world settings~\cite{yang2025causalmob}. Ideally, a model should use structured information during training but still perform well when it is missing at inference time.
(2) Networks have graph topology, so event impacts propagate along edges. In Figure~\ref{fig:teaser}, two geographically close locations can still be topologically disconnected, yet models built on classical denoiser backbones~\cite{zhang2025chattraffic,ge2025t2s} often fail to capture such propagation.
Moreover, this issue is rarely considered in evaluation, leaving topology-aware metrics largely missing.
Most widely used evaluation metrics~\cite{yuan2024diffusionts,hosseini2025quality} give the same score even when the node orders are permuted.

% Diffusion models typically inject conditional information either through structured conditions that are spatially aligned with the output via auxiliary control branches or gating mechanisms~\cite{zhang2023controlnet}, or through unstructured conditions such as text that can convey richer and more detailed information, influencing generation primarily via cross-attention~\cite{rombach2022ldm} or guidance~\cite{dhariwal2021diffusion}.
% Although some works~\cite{draxler2025transformers} attempt to encode them as special tokens concatenated with unstructured conditions, such tokens have limited expressive power and struggle with complex spatio-temporal tasks.

To address these challenges, we propose \task{} (\underline{\textbf{Net}}work \underline{\textbf{Ev}}ent \underline{\textbf{Ev}}olution), a generative simulator for network event evolution that jointly incorporates multi-structural event cues while preserving network topology during generation. 
We introduce a structure-guided masking objective that encourages the encoder to learn event context informed by structured attributes. 
Building on the resulting latent space, \task{} generates future network evolution with a topology-aware denoising process, while AdaLN~\cite{ge2025t2s} enables flexible modulation by unstructured event descriptions. 
Specifically, topology awareness is achieved by downsampling and then upsampling the graph during denoising, so information aggregates and propagates along network connections at each stage.
Importantly, after training, \task{} can run using only natural-language event inputs at inference time, offering greater flexibility.
To support learning and evaluation at scale, we introduce \task{}-6.5M, a multimodal benchmark with over 6.5 million aligned event and network evolution pairs spanning four metropolitan road networks, avoiding reliance on LLM-produced annotations~\cite{gu2025verbalts,ge2025t2s} or costly manual labeling.
We further propose a topology-sensitive metric, JL-MMD, inspired by the Johnson–Lindenstrauss lemma, to efficiently assess whether generated dynamics exhibit realistic graph propagation patterns. 
Our contributions are summarized as follows:
\begin{itemize}
    \item We formulate a new and challenging problem: simulating the evoluation of network states under textual event conditions.
    \item We propose \task{}, a generative simulator that combines structure-guided masked pretraining with a topology-aware denoiser, which demonstrates superior performance with generalization ability in extensive experiments.
    \item We construct \task{}-6.5M, a large-scale multimodal benchmark of aligned event text and network traffic data.
    \item We introduce JL-MMD, a scalable metric based on Johnson-Lindenstrauss random projections, to evaluate topological fidelity in generated network dynamics.
\end{itemize}

%% file: content/3_Preliminaries.tex
\input{tables/dataset.tex}

\section{Preliminaries}
\label{sec:preliminaries}

\subsection{Problem Formulation}

We model a network as a graph $G=(V,\mathbf{A})$, where $V$ is a set of $N$ sensor nodes and $\mathbf{A}\in\mathbb{R}^{N\times N}$ is the adjacency matrix representing the network topology. Let $\mathbf{X}\in\mathbb{R}^{T\times N}$ denote a sequence of network snapshots (e.g., traffic flow) over $T$ time steps.
Each event is represented by a set of multi-structural attributes $E={E_i}_{i=1}^{M}$, including a textual description $E_d$ and structured properties.
Given these event conditions as initial state, the task is to simulate realistic future states $\hat{\mathbf{X}}$ that reflect how an event evolves and propagates over the network topology. Formally,
\begin{equation}
E={E_i}_{i=1}^{M} \xrightarrow{G} \hat{\mathbf{X}}\in\mathbb{R}^{T\times N},
\label{eq:problem}
\end{equation}
In this work, we instantiate $E=(E_d, E_n, E_t)$, corresponding to the textual description for a network event, spatial context (i.e., indices of the event node and impacted nodes in the egocentric subgraph), and temporal context, respectively. %(see Section~\ref{sec:data_construction} for details).

\subsection{\task{}-6.5M Benchmark}
\label{sec:data_construction}
We construct a new multimodal benchmark by integrating traffic signal observations from LargeST~\cite{liuLargeSTBenchmarkDataset} with event records from LSTW~\cite{lstw_pattern_discovery}, covering 4 road networks named San Diego (SD), Greater Bay Area (GBA), Greater Los Angeles (GLA), and California (CA).
Traditionally, event research and network traffic have been studied in separate directions~\cite{gou2025traffident}, so there has been limited effort to align the data. We thereby conduct a careful rule-based matching between event locations and traffic sensors, described as follows.

\subsubsection{Event-traffic Alignment}
We consider two major types of events on networks, namely incidents and weathers. 

\noindent\textbf{Incident Alignment.}
Traffic incidents are matched to sensors using spatial proximity and directional consistency: (1) project sensor and incident coordinates to EPSG:3857; (2) create a 100m buffer around each sensor and retrieve incidents within the buffer; (3) extract highway directions (N/S/E/W) from incident street names and discard pairs with inconsistent directions; and (4) align incident durations to 5-minute sensor intervals.

\noindent\textbf{Weather Alignment.} Weather events are reported at meteorological stations with broader spatial coverages. 
We use Voronoi tessellation to link sensors to stations: (1) build Voronoi cells from station locations; (2) assign each traffic sensor to the station whose cell contains it; and (3) let sensors inherit the weather conditions of their assigned station. 
This deterministic matching strategy makes both incident and weather alignment reproducible across cities.

\subsubsection{Event Descriptions}
LSTW~\cite{lstw_pattern_discovery} provides textual descriptions for traffic incidents, but not for weather events. 
To fill this gap, we generate location descriptions for weather stations using a large language model (LLM). 
Because many stations are not tied to well-known place names, directly querying the Maps API often yields coordinate-like identifiers such as plus codes, e.g., ``P27Q+MCM'', which are highly structured and do not match our goal of using natural language as the primary unstructured condition. 
Instead, we use Google Maps to obtain representative views, and manually adjust the map position to best distinguish each station, and capture screenshots. 
We then feed these screenshots to the LLM and prompt it to produce natural-language descriptions.
Table~\ref{tab:sample_records} provides examples of aligned incident and weather records, together with the resulting textual condition used by our benchmark.
\input{tables/sample_records_compact}

\subsubsection{Multi-structual Sample Organization}
% We select $E_n$ and $E_t$ as the structured attributes because they are the only event properties that cannot be adequately represented through text.
Given the descriptions of incidents and weather events, attributes like severity, road address, direction, and event type can naturally by expressed in the unstructured part $E_d$.
In contrast, $E_n$ and $E_t$ are the textual parts for presenting the structured spatial contexts and temporal attributes of events, which are beyond the expressiveness of text in a succinct way. The former consists of a target node where an event takes place and its egocentric subgraph that the event would impact, while the latter describes time-related information (e.g., spans) in precise numerical values (sometimes in different formats).
% In contrast, node indices encode positions in the graph topology with no natural textual mapping, and temporal spans require precise numerical values that text descriptions only convey approximately. 
% Therefore, $E_n$ and $E_t$ are the sufficient structured attributes to the textual condition.

We yield the first large-scale benchmark \task{}-6.5M of aligned network event-traffic data. As summarized in Table~\ref{tab:dataset_stats}, there are over 6.5 million samples, taking approximately 12~TB of storage. Each sample consisting of: (1) \textit{Textual description} $\mathbf{E}_d$, the event description encoded by a text encoder; (2) \textit{Network snapshots} $\mathbf{X} \in \mathbb{R}^{T \times N}$, the subsequent time steps (e.g., $T=96$, or 8 hours at a 5-minute interval) following the event; (3) \textit{Spatial context} $E_n$, the target node of the event and its n-hop (e.g., 2-hop) neighbors over the network; (4) \textit{Temporal context} $E_t$, the event’s start and end timestamps relative to the snapshots.
%\task{}-6.5M is the first benchmark of aligned network event evolution pairs (over 6.5 million samples, taking approximately 12~TB of storage). Table~\ref{tab:dataset_stats} summarizes its statistics, including the number of samples.

%% file: tables/dataset.tex
\begin{table*}[htbp]
    \centering
    \caption{Summary of \task{}-6.5M benchmark (*no detailed incident types are reported for GLA and GBA in LSTW~\cite{lstw_pattern_discovery})}
    \label{tab:dataset_stats}
    \setlength{\tabcolsep}{1.5pt}
    \resizebox{\textwidth}{!}{
    \begin{tabular}{l|r|rrrrrrr|rrrrrr}
    \toprule
    \toprule
    \multirow{2}{*}{\textbf{Subset}} & \multirow{2}{*}{\textbf{Total}} & \multicolumn{7}{c|}{\textbf{Incident Events}} & \multicolumn{6}{c}{\textbf{Weather Events}} \\
    \cmidrule(lr){3-9} \cmidrule(lr){10-15}
     & & \textbf{Accident} & \textbf{Broken-Vehicle} & \textbf{Congestion} & \textbf{Construction} & \textbf{Event} & \textbf{Lane-Blocked} & \textbf{Flow-Incident} & \textbf{Severe-Cold} & \textbf{Fog} & \textbf{Hail} & \textbf{Rain} & \textbf{Snow} & \textbf{Storm} \\
    \midrule
    \textbf{SD}  & 836,694   & 5,804  & 2,049   & 33,933  & 2,676  & 2   & 3,847  & 3,536  & 440,186   & 740    & 82,308  & 265,359   & 810    & 561   \\
    \textbf{GLA} & 654,420   & *51,086 & 0       & 0       & 0      & 0   & 0      & 0      & 237,202   & 1,536  & 243,411 & 125,394   & 1,328  & 99    \\
    \textbf{GBA} & 1,201,201 & *22,639 & 0       & 0       & 0      & 0   & 0      & 0      & 329,502   & 3,442  & 383,211 & 462,014   & 3,888  & 83    \\
    \textbf{CA}  & 3,824,877 & 33,427 & 9,709   & 189,586 & 11,195 & 33  & 23,216 & 15,919 & 1,252,645 & 3,252  & 338,278 & 1,965,992 & 15,980 & 8,145 \\
    \midrule
    \textbf{Total} & \textbf{6,517,192} & \textbf{112,956} & \textbf{11,758} & \textbf{223,519} & \textbf{13,871} & \textbf{35} & \textbf{27,063} & \textbf{19,455} & \textbf{2,259,535} & \textbf{8,970} & \textbf{1,047,208} & \textbf{2,818,759} & \textbf{22,006} & \textbf{8,888} \\
    \bottomrule
    \bottomrule
    \end{tabular}
    }
\end{table*}

%% file: tables/sample_records_compact.tex
\begin{table}[t]
    \centering
    \caption{Sample event records in \task{}-6.5M}
    \label{tab:sample_records}
    \footnotesize
    \setlength{\tabcolsep}{2.0pt}
    \begin{tabularx}{\columnwidth}{>{\raggedright\arraybackslash}p{0.1\columnwidth}>{\raggedright\arraybackslash}p{0.2\columnwidth}>{\raggedright\arraybackslash}X}
        \toprule
        \toprule
        \textbf{Source} & \textbf{Type / Place} & \textbf{Condition text} \\
        \midrule
        Incident & Congestion / I5-S & Severe delays of ten minutes on San Diego Fwy Southbound in Camp Pendleton South. Average speed 20 mph. \\
        Incident & Lane-blocked / CA-15 & Lane blocked due to debris on road on CA-15 at Exit 6A Adams Ave. \\
        Weather & Light rain / Mission Bay & Light rain weather event occurred at the intersection of Mission Bay Drive and Ingraham Street, near SeaWorld San Diego. \\
        Weather & Severe storm / Calabasas & Severe storm weather event occurred at the intersection of Buena Vista Drive and South Drive, near Brothers Market, with Calabasas Road to the northeast and open space to the south. \\
        \bottomrule
        \bottomrule
    \end{tabularx}
\end{table}

%% file: content/4_Methodology.tex
\section{Methodology}
\label{sec:method}
\begin{figure*}[h]
    \centering
    \includegraphics[width=1\textwidth]{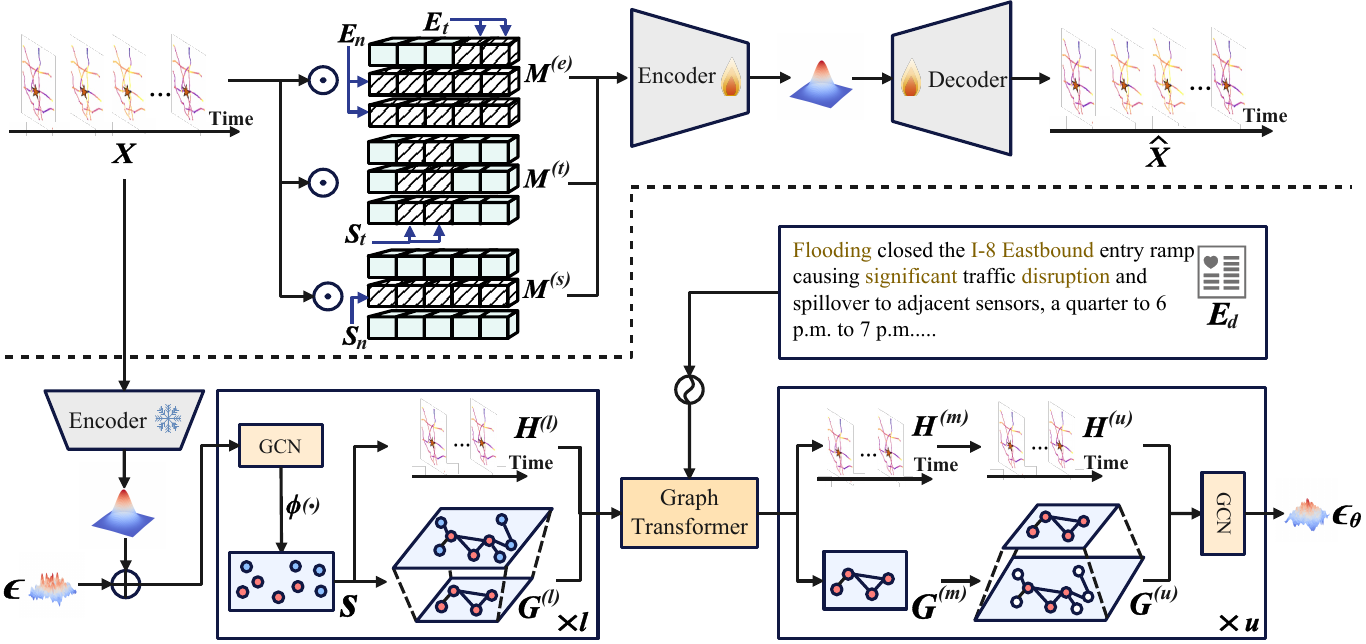}
    \caption{Net-Ev$^2$ framework: the upper illustrates the structure-guided masked pre-training; the lower depicts the topology-aware diffusion process.}
    \Description{The framework first masks network snapshots using event, temporal, and spatial structure before encoding and decoding them during pre-training. The diffusion stage combines a latent network state, event text embedding, graph transformer, and graph downsampling and upsampling modules to generate future network evolution.}
    \label{fig:framework} 
\end{figure*}
In this section, we present \method{}, as illustrated in Figure~\ref{fig:framework}.
The input, the sequence of network snapshots $\mathbf{X}$, is fed into the encoder with both spatio-temporal masks and structure-guided masks, following a Masked Autoencoder (MAE) design. 
During training, we add noise to the hidden representations, and a topology-aware denoiser removes it using paired pooling and unpooling operations to preserve network topology.
Unstructured event conditions are injected at the bottleneck via AdaLN~\cite{ge2025t2s}. 
During inference, we sample latent representations from noise and decode them using the pre-trained decoder.

\subsection{Structure-guided MAE}

The Structure-guided MAE compresses traffic observations into latent space where diffusion operates efficiently. 
During training, it leverages multi-structural event information to guide masking strategies, enabling more targeted representation learning. 
Compared with~\cite{draxler2025transformers} which attempts to encode them as special tokens, this approach offers more powerful representation capability.

\subsubsection{Mask Generator}
In this part, we introduce two masking strategies. Since structure-guided masks are fixed across inputs, we additionally use a fully random spatiotemporal mask for diversity~\cite{10.24963/ijcai.2024/442}. 
The three masked passes share the same encoder--decoder parameters, each producing an independent posterior and reconstruction. We fuse their outputs via residual addition, forcing one latent space to remain robust to random temporal, random spatial, and event-guided missing patterns.
The mask generator produces spatial and temporal masking patterns to encourage robust representations:

\noindent\textbf{Spatio-temporal Masking} We construct binary masks $\mathbf{M}^{(t)}, \mathbf{M}^{(s)} \in \{0, 1\}^{T \times N}$ by randomly selecting subsets $\mathcal{T} \subset \{1, \ldots, T\}$ and $\mathcal{S} \subset \{1, \ldots, N\}$ with $|\mathcal{T}| = \lfloor \rho \cdot T \rfloor$ and $|\mathcal{S}| = \lfloor \rho \cdot N \rfloor$, then setting $\mathbf{M}^{(t)}_{i,:} = 0$ for $i \in \mathcal{T}$ and $\mathbf{M}^{(s)}_{:,j} = 0$ for $j \in \mathcal{S}$. This masks the entire time steps and sensor nodes, respectively.

\noindent\textbf{Structure-guided Masking}
This mask is constructed by masking the affected nodes $E_n$ and the corresponding time windows determined by $E_t$:
\begin{equation}
    \mathbf{M}^{(e)}_{:,j} = 0 \text{ for } j \in E_n, \quad \mathbf{M}^{(e)}_{i,:} = 0 \text{ for } i < \max(E_t)
\end{equation}
Here, $E_t$ denotes the event temporal span, and $\max(E_t)$ selects its last time step. Since the generated data is defined to begin after the event starts, using $i < \max(E_t)$ directly masks the event duration.

\subsubsection{Variational Encoder-decoder Architecture}
We adopt a variational formulation~\cite{jordan1998variational} to improve generalizability. 
The encoder maps the masked input into a compact latent representation, and the decoder reconstructs the original signal from this latent space. Specifically, the encoder first projects the input into the latent space and then applies Transformer layers:
\begin{equation}
    [\boldsymbol{\mu}, \log \boldsymbol{\sigma}^2] = \text{TransformerEncoder}(\text{Linear}(\mathbf{X} \odot \mathbf{M}))
\end{equation}
Then we sample $\mathbf{z} = \boldsymbol{\mu} + \boldsymbol{\sigma} \odot \boldsymbol{\xi}$ with $\boldsymbol{\xi} \sim \mathcal{N}(0, \mathbf{I})$ during training, and use $\mathbf{z} = \boldsymbol{\mu}$ during inference.
The decoder mirrors the encoder structure, mapping $\mathbf{z}$ back to the original space $\hat{\mathbf{X}} \in \mathbb{R}^{T \times N}$.

\subsubsection{Training Objective}

Finally, the masked inputs are fed into the shared variational encoder--decoder through three parallel passes under $\mathbf{M}^{(t)}$, $\mathbf{M}^{(s)}$, and $\mathbf{M}^{(e)}$. Their outputs are fused by residual addition, $\hat{\mathbf{X}}=\hat{\mathbf{X}}^{(t)}+\hat{\mathbf{X}}^{(s)}+\hat{\mathbf{X}}^{(e)}$, and the model is optimized with a loss that trades off reconstruction accuracy against KL-divergence regularization~\cite{yu2013kl} over all latent posteriors:
\begin{equation}
    \mathcal{L}_{MAE} = \|\mathbf{X} - \hat{\mathbf{X}}\|_2^2 + w \sum_{r \in \{t,s,e\}} D_{KL}(q_r(\mathbf{z}|\mathbf{X}, \mathbf{M}^{(r)}) \| p(\mathbf{z}))
\end{equation}
where $p(\mathbf{z}) = \mathcal{N}(0, \mathbf{I})$ is the prior distribution, and each KL term $D_{KL} = \frac{1}{2}\sum(\boldsymbol{\mu}^2 + \boldsymbol{\sigma}^2 - \log\boldsymbol{\sigma}^2 - 1)$ regularizes the latent space to smooth for subsequent diffusion.

\subsection{Topology-aware Diffusion Model}

Topology-aware diffusion model operating entirely in the latent space, maintaining the graph hierarchy across all denoising steps.

\subsubsection{Diffusion Process}

We adopt the DDPM~\cite{ho2020denoising} method in the latent space. 
Let $\mathbf{z}_0$ denote the latent representation from the MAE. 
The forward process adds Gaussian noise $q(\mathbf{z}_t | \mathbf{z}_0) = \mathcal{N}(\mathbf{z}_t; \sqrt{\bar{\alpha}_t}\mathbf{z}_0, (1-\bar{\alpha}_t)\mathbf{I})$, where $\bar{\alpha}_t = \prod_{r=1}^{t} \alpha_r$ with noise schedule $\{\beta_t\}$. 
The reverse process learns to denoise conditioned on $\mathbf{c}$:
\begin{equation}
    p_\theta(\mathbf{z}_{t-1} | \mathbf{z}_t, \mathbf{c}) = \mathcal{N}(\mathbf{z}_{t-1}; \boldsymbol{\mu}_\theta(\mathbf{z}_t, t, \mathbf{c}), \tilde{\beta}_t \mathbf{I})
\end{equation}

\subsubsection{Graph U-Net Architecture}
We incorporate event conditions into the Graph U-Net to preserve the network topology, since each Graph U-Net block propagates information and encourages the model to learn inter-node relationships.
The Graph U-Net takes $(\mathbf{z}_t, t, \mathbf{c})$ as inputs and predicts the noise $\boldsymbol{\epsilon}_\theta$. It consists of an encoder-bottleneck-decoder structure with $L$ levels and skip connections: the encoder progressively pools the graph from $G^{(0)}$ to $G^{(L)}$; the bottleneck injects conditioning; the decoder unpools back to the original resolution.

\noindent\textbf{Graph Pooling} Learnable pooling is introduced to build hierarchical representations while preserving salient topology.
A projection layer computes node importance scores $\boldsymbol{\pi}=\operatorname{sigmoid}(\text{Linear}(\mathbf{H}^{(l)}))\in\mathbb{R}^{N_l}$, and the top-$k$ nodes are selected as $\mathcal{I}=\text{top-}k(\boldsymbol{\pi})$.
The pooled features are then formed as $\mathbf{H}^{(l+1)}=\mathbf{H}^{(l)}{\mathcal{I}}\odot \boldsymbol{\pi}{\mathcal{I}}$, where $\boldsymbol{\pi}_{\mathcal{I}}$ serves as a gating mechanism that emphasizes informative nodes. The adjacency matrix is coarsened accordingly to maintain connectivity.

\noindent\textbf{Graph Transformer Block} At the bottleneck, this block injects event conditioning through AdaLN~\cite{ge2025t2s}. Unlike standard LayerNorm with fixed affine parameters, AdaLN derives the scale $\boldsymbol{\gamma}$, shift $\boldsymbol{\delta}$, and gate $\boldsymbol{\lambda}$ from the conditioning vector $\mathbf{c}$:
\begin{equation}
    [\boldsymbol{\gamma}, \boldsymbol{\delta}, \boldsymbol{\lambda}] = \text{Linear}(\mathbf{c}), \quad \text{AdaLN}(\mathbf{H}, \mathbf{c}) = \boldsymbol{\gamma} \odot \text{LayerNorm}(\mathbf{H}) + \boldsymbol{\delta}
\end{equation}
Each sub-layer applies AdaLN before its operation and uses the corresponding gate $\boldsymbol{\lambda}$ to scale the residual, where $\text{GCN}(\cdot)$ denotes a Graph Convolution Network:
\begin{align}
    \mathbf{H}_g &= \mathbf{H}^{(L)} + \boldsymbol{\lambda}_g \odot \text{GCN}(\text{AdaLN}(\mathbf{H}^{(L)}, \mathbf{c}), \mathbf{A}^{(L)}) \\
    \mathbf{H}_a &= \mathbf{H}_g + \boldsymbol{\lambda}_a \odot \text{Attention}(\text{AdaLN}(\mathbf{H}_g, \mathbf{c})) \\
    \mathbf{H}_m &= \mathbf{H}_a + \boldsymbol{\lambda}_m \odot \text{MLP}(\text{AdaLN}(\mathbf{H}_a, \mathbf{c}))
\end{align}
where Attention~\cite{10.5555/3295222.3295349} corresponds to the standard Transformer self-attention module.

\noindent\textbf{Graph Unpooling} The decoder restores graph resolution by placing pooled features back to original positions via $(\mathbf{H}^{(u-1)}, G^{(u-1)}) = \text{unpool}(\mathbf{H}^{(u)}, \boldsymbol{\pi}^{(u)}, \mathbf{A}^{(u-1)})$, with GCN layers propagating information to unselected nodes. A residual connection ensures gradient flow, and a final linear layer projects $\mathbf{H}_{out}$ to the predicted noise $\boldsymbol{\epsilon}_\theta$.

\subsubsection{Conditioning Mechanism}
We condition on the diffusion timestep and event text embedding $\mathbf{E}_d$. Following the existing practice~\cite{ge2025t2s,zhang2025chattraffic}, timesteps are encoded using sinusoidal positional embeddings, and the text embedding is obtained via pre-trained BERT~\cite{devlin-etal-2019-bert}. These are combined to form the conditioning vector $\mathbf{c}$. Structured attributes guide representation learning through the masking objective, but are not required by the denoiser at inference, which keeps scenario simulation accessible from natural-language event descriptions alone.

\subsubsection{Training Objective}

The denoiser is trained to predict the noise added at each diffusion step:
\begin{equation}
    \mathcal{L}_{diff} = \mathbb{E}_{t, \mathbf{z}_0, \boldsymbol{\epsilon}} \left[ \|\boldsymbol{\epsilon} - \boldsymbol{\epsilon}_\theta(\mathbf{z}_t, t, \mathbf{c})\|_2^2 \right]
\end{equation}

\subsection{Inference}

During inference, we start from Gaussian noise $\mathbf{z}_{T_{diff}} \sim \mathcal{N}(0, \mathbf{I})$ and iteratively denoise:
\begin{equation}
    \mathbf{z}_{t-1} = \frac{1}{\sqrt{\alpha_t}} \left( \mathbf{z}_t - \frac{1-\alpha_t}{\sqrt{1-\bar{\alpha}_t}} \boldsymbol{\epsilon}_\theta(\mathbf{z}_t, t, \mathbf{c}) \right) + \sqrt{\tilde{\beta}_t} \boldsymbol{\epsilon}
\end{equation}
where $\boldsymbol{\epsilon} \sim \mathcal{N}(0, \mathbf{I})$ for $t > 1$ and $\boldsymbol{\epsilon} = 0$ for $t = 1$. The final $\mathbf{z}_0$ is decoded to obtain $\hat{\mathbf{X}}$.

%% file: content/5_Experiments.tex
\input{tables/overall}

\input{tables/ablation}

\section{Experiments}
% This paragraph can be commented out if necessary

% This section presents comprehensive experiments on \method{}. Section~\ref{sec:overall} presents the overall performance comparison against baseline methods. We then analyze the contribution of each component through ablation studies in Section~\ref{sec:ablation} and examine generalization capabilities in Section~\ref{sec:generalization}. Section~\ref{sec:hyperparam} investigates the sensitivity to key hyper-parameters, followed by efficiency analysis in Section~\ref{sec:efficiency}. Finally, Section~\ref{sec:case} provides qualitative case studies to illustrate the controllability and fidelity of generated simulations by \method{}.

% to demonstrate the practical effectiveness of our approach.

\subsection{Experimental Setup}
\label{sec:setup}

\noindent\textbf{Implementation Details.}
\method{} is implemented using PyTorch 2.3.1+cu121 with Python 3.10.18, and trained on a server equipped with four NVIDIA RTX A800 GPUs (80GB memory each). We use the Adam optimizer with an initial learning rate of 0.0001. To avoid leaking future events, samples are split chronologically by event order, with the first 80\% used for training and the remaining 20\% used for testing. During inference, all models perform 100 sampling iterations, and we report average scores of each metric over five runs with different random seeds. The input window size is set to 96 time steps for all models. All experiments are conducted on the multimodal benchmark \task{}-6.5M introduced in Section~\ref{sec:data_construction}. Full year of 2017 data is used for SD, GLA and GBA, while January 2017 is used for CA due to the large scale of its network.

%\noindent\textbf{Datasets.}
%We evaluate on the multimodal dataset constructed in Section~\ref{sec:data_construction}, which integrates traffic observations with event records across four metropolitan networks, namely SD, GLA, GBA, and CA. For SD, GLA, and GBA, we use the full year of 2017 data for training and testing. For CA, due to its large scale, we only use the first month of 2017.

\noindent\textbf{Baselines.}
We evaluated \method{} with the following baselines: %state-of-the-art \task{} methods.
\begin{itemize}
    \item \textbf{Diffusion-TS}~\cite{yuan2024diffusionts} employs an encoder-decoder transformer with disentangled temporal representations, incorporating trend and seasonality decomposition to capture semantic meaning. It directly learns from raw time series without Variational Autoencoder (VAE) compression.
    \item \textbf{VerbalTS}~\cite{gu2025verbalts} generates time series from unstructured textual descriptions through a multi-focal alignment and generation framework. It also directly operates on raw time series without VAE.
    \item \textbf{ChatTraffic}~\cite{zhang2025chattraffic} transforms traffic data into image representations and applies diffusion models with GCNs to capture spatial correlations.
    \item \textbf{T2S}~\cite{ge2025t2s} bridges natural language and time series using Flow Matching with Diffusion Transformer (DiT). It employs a length-adaptive VAE to encode variable-length sequences into consistent latent embeddings, enabling generation of arbitrary lengths.
\end{itemize}

\noindent\textbf{Evaluation Metrics.}
We employ five complementary metrics to evaluate the generation quality. \textbf{ED}, Euclidean Distance, directly measures the point-wise distance between generated and ground truth sequences. \textbf{C-FID}, Context Fr\'echet Inception Distance, computes the semantic distance in latent space using TS2Vec~\cite{ts2vec} as the encoder. \textbf{MRR@5}, Mean Reciprocal Rank, calculates the reciprocal of the rank at which the similarity score first exceeds a threshold across five runs. \textbf{CTTP}, Contrastive Time series Text Pretraining~\cite{gu2025verbalts}, is inspired by CLIP-style cross-modal retrieval~\cite{DBLP:journals/corr/abs-2103-00020}; it maps network dynamics and text conditions into a shared embedding space and reports retrieval accuracy, where higher values indicate stronger condition-following. \textbf{JL-MMD} is the topology-aware metric we propose to assess topological fidelity. Designing such a metric is challenging, as topology requires spatial sensitivity to assess distribution alignment, and each sample spans multiple time steps with distinct graphs, limiting large-scale evaluation. Recent work~\cite{hosseini2025quality} introduces the Johnson-Lindenstrauss lemma for dynamic graph evaluation, but only supports single sample pair analysis. We extend it to batch evaluation through Maximum Mean Discrepancy (MMD). Given batches of real graph embeddings $\{\mathbf{h}_{G_i}^{\text{real}}\}_{i=1}^{n_1}$ and generated embeddings $\{\mathbf{h}_{G_j}^{\text{gen}}\}_{j=1}^{n_2}$, JL-MMD is defined:
\begin{equation}
    \text{JL-MMD} = \left\|\frac{1}{n_1}\sum_{i=1}^{n_1} \boldsymbol{\Phi}\mathbf{h}_{G_i}^{\text{real}} - \frac{1}{n_2}\sum_{j=1}^{n_2} \boldsymbol{\Phi}\mathbf{h}_{G_j}^{\text{gen}}\right\|_2^2
\label{eq:jl-mmd}
\end{equation}
where $\boldsymbol{\Phi} \in \mathbb{R}^{k \times d}$ is a random projection matrix that preserves pairwise distances. All main experiments use the linear kernel for computational efficiency ($O(nd)$ vs $O(n^2d)$ for RBF kernels). Appendix~\ref{app:jlmmd-validation} provides the synthetic metric validation and the RBF-kernel sanity check.

\subsection{Performance Evaluation}
\label{sec:overall}

We compare \method{} with baseline models across all four datasets and report results for three horizons of 24, 48, and 96 time steps. According to Table~\ref{tab:overall}, we summarize findings as follows:
\begin{itemize}
    \item Our model achieves the best performance across all metrics, particularly on JL-MMD which measures graph structural similarity. Baseline models achieve varying results on other metrics. However, their JL-MMD scores are strikingly similar, suggesting that JL-MMD is particularly sensitive to topological structure and hard to improve without explicit graph modeling.
    \item Considering different generation horizons, distance-based metrics (both Euclidean distance and latent space distance) tend to deteriorate as the generation length increases due to error accumulation. In contrast, MRR@5 improves with longer horizons since it measures relative ranking rather than absolute values.
    \item The simulation task becomes easier as the generation horizon increases. Since the longest setting (96 steps) spans 8 hours and network events rarely have such long-lasting effects, long-horizon generation is less challenging.
    \item Regarding network scale, ED goes up as the network scale grows, while C-FID remains relatively scale-invariant as it computes similarity in latent space. MRR@5 also benefits from larger datasets since they contain more easily learned node patterns, resulting in higher similarity scores. JL-MMD does not exhibit a direct correlation with graph size but rather reflects the complexity of topological structures, as smaller graphs like SD may still possess intricate topologies.
    \item Comparing different model architectures, VAE-based models consistently outperform non-VAE alternatives. Although non-VAE models are designed for time series, networked data contain substantial information and require appropriate compression to avoid dimensional explosion.
\end{itemize}

\subsection{Condition-following Analysis}
\label{sec:cttp}

While ED, C-FID, and JL-MMD evaluate distributional and topological fidelity, they do not directly measure whether the simulated network dynamics follow the given text conditions. We therefore report CTTP accuracy in Table~\ref{tab:cttp}. 
\input{tables/cttp}
\method{} consistently achieves the highest CTTP scores across SD, GLA, and GBA, showing that the generated dynamics are more aligned with event descriptions than those produced by text-to-series baselines. The ground-truth scores provide an approximate upper bound and also quantify data alignment quality; generated results should not exceed this bound unless the model overfits the retrieval evaluator. CTTP is also sensitive to network scale, so we use it as a condition-following complement rather than a replacement for JL-MMD.

\subsection{Ablation Study}
\label{sec:ablation}

To evaluate the contribution of each component in \method{}, we adopt six variants: \textbf{W/o $\mathbf{M}^{(e)}$} removes the structure-guided mask and only uses text as the condition during training; \textbf{W/o $\mathbf{M}^{(t,s)}$} removes the random spatio-temporal masks and only uses structure-guided masks, lacking randomness in the masking process; \textbf{W/o $\mathbf{E}_d$} is an unconditional variant that removes text conditioning during inference; \textbf{W/o $\mathbf{E}_d^{(w)}$} removes weather-related text descriptions; \textbf{W/o $\mathbf{E}_d^{(i)}$} removes incident-related text descriptions; \textbf{W/o Graph U-Net} replaces the Graph U-Net denoiser with a simple DiT that lacks explicit topological modeling.

As shown in Table~\ref{tab:ablation}, the full model consistently outperforms all variants. Several key observations emerge:
\input{tables/generalization}

\begin{itemize}
    \item Removing the structure-guided mask $\mathbf{M}^{(e)}$ causes more performance degradation than removing the spatio-temporal masks $\mathbf{M}^{(t,s)}$, indicating that structure-guided masking is more critical for learning event evolution patterns on networks. Furthermore, the performance after removing $\mathbf{M}^{(e)}$ drops to a level comparable to ChatTraffic and worse than T2S, which is understandable as T2S employs a specially-designed encoder for taking in variable-length inputs.
    \item Variants without the MAE components exhibit larger performance drops than data-level variants (i.e., W/o $\mathbf{E}_d$, W/o $\mathbf{E}_d^{(w)}$, W/o $\mathbf{E}_d^{(i)}$, implying MAE is essential for learning quality representations in the latent space.
    \item Both weather and incident text descriptions contribute to generation quality, as the conditional variants outperform the unconditional variant W/o $\mathbf{E}_d$, demonstrating that \method{} can leverage various conditions effectively.
    \item Replacing Graph U-Net with DiT results in relatively small changes in ED and other metrics, given DiT is also a powerful denoiser. However, doing so causes JL-MMD to degrade substantially, confirming the capability of Graph U-Net to capture hierarchical structure of the network topology.
\end{itemize}

\subsection{Generalization Studies}
\label{sec:generalization}

\subsubsection{Generalization across years and networks}

We also conduct transfer experiments to evaluate the generalization ability of \method{}, including cross-network transfer (i.e., from small to large networks, from large to small ones) and cross-temporal transfer (i.e., from one year to future years). 
Since the transfer changes the network size, an adapter is added for fair comparison. The adapter is a single linear layer that projects target-domain nodes into the source-domain node space, and all model inputs and losses are computed after this mapping.
We report results for both fully fine-tuned and fewshot settings with 10\% of the target data in Table~\ref{tab:generalization}.

Under these challenging transfer settings, most models show lower simulation fidelity, yet \method{} successfully transfers graph structure knowledge and maintains acceptable JL-MMD scores by significantly outperforming all baselines on this topology-sensitive metric. In contrast, Diffusion-TS shows weak transfer capability with substantially higher ED and C-FID values, as it directly learns from raw time series without latent compression (by VAE). Similarly, ChatTraffic suffers from poor transferability, likely because its image-based representation is sensitive to changes in network topology. Notably, fewshot training achieves comparable or sometimes even better performance than full finetuning, suggesting that freezing most parameters helps prevent overfitting when the training data is large. We also observe that transferring from larger networks to smaller ones (e.g., GBA to GLA) in general yields slightly better performance than the reverse direction, as models trained on larger networks may have learnt diverser and more representative patterns.

\input{tables/psme}
% consider moving this subsection to appendix - to save some space for better related work

\subsubsection{Generalization beyond \task{}-6.5M}
\label{sec:psme}
To test whether \method{} applies beyond road networks, we further evaluate it on the Power System Multi-source Events Dataset (PSME)~\cite{tj1p-4y20-26}. PSME is generated from the IEEE 34 Node Test Feeder through electromagnetic transient simulation and provides graph-structured PMU recordings sampled at 120 Hz. We convert each label structure into a concise textual condition, and align it with the corresponding PMU trajectory.

As shown in Table~\ref{tab:psme}, \method{} achieves the best ED, C-FID, and JL-MMD scores over 96-step simulations. This result indicates that topology-aware diffusion extends beyond traffic networks, as the effects of events also propagate through the structured connectivity of power systems.
% This result suggests that topology-aware diffusion works beyond traffic networks, as event impacts also propagate through structured connectivity in power systems.

\subsection{Hyper-parameter Study}
\label{sec:hyperparam}

\begin{figure}[ht]
    \centering
    \includegraphics[width=0.48\textwidth]{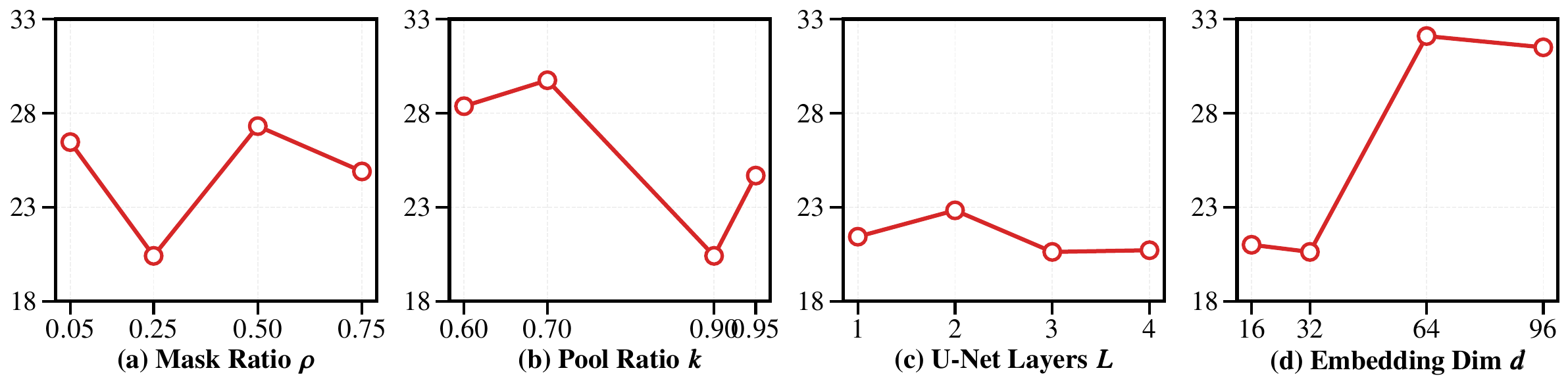}
    \caption{Hyper-parameter study on SD}
    \Description{Four line plots report Euclidean distance on the SD dataset as mask ratio, pool ratio, number of U-Net layers, and embedding dimension vary. The curves show the lowest error around mask ratio 0.25, pool ratio 0.90, three to four U-Net layers, and smaller embedding dimensions.}
    \label{fig:para_sd_ed}
\end{figure}

We study four key hyper-parameters on SD in Figure~\ref{fig:para_sd_ed}, using ED as the evaluation metric.
\begin{itemize}
    \item \textbf{Mask Ratio $\rho$}: $\rho \in \{0.05, 0.25, 0.50, 0.75\}$. The best value is 0.25, where ED is minimized. Smaller $\rho$ provides insufficient masking for robust representation learning, while larger $\rho$ masks too much information and hinders reconstruction.
    \item \textbf{Pool Ratio $k$}: $k \in \{0.60, 0.70, 0.90, 0.95\}$ in Graph U-Net. The best value is 0.90. Smaller $k$ retains too many nodes and introduces computational redundancy, while $k=0.95$ discards too many important nodes and loses critical structural information.
    \item \textbf{Number of U-Net Layers $L$}: $L \in \{1, 2, 3, 4\}$. Performance improves with more layers, with 3--4 layers performing best. Fewer layers have insufficient capacity to capture complex hierarchical graph structures.
    \item \textbf{Embedding Dimension $d$}: $d \in \{16, 32, 64, 96\}$. Smaller dimensions (16--32) perform better. Larger dimensions increase model capacity and lead to overfitting relative to the task complexity.
\end{itemize}

\subsection{Efficiency Study}
\label{sec:efficiency}
\input{tables/efficiency_tables}

We compare the training efficiency of the denoising stage across all methods on a single NVIDIA RTX A800 GPU, as shown in Table~\ref{tab:efficiency}. \method{} achieves the best efficiency across all datasets. This is because Graph U-Net leverages GCN operations that scale linearly with the number of edges, avoiding the quadratic complexity of attention mechanisms in standard DiT. In contrast, T2S employs DiT as the denoiser, resulting in significantly higher computational cost, especially on large-scale networks like CA. Furthermore, Diffusion-TS and VerbalTS operate directly on raw time series without latent space compression, leading to higher dimensionality during denoising and consequently lower efficiency compared to VAE-based methods.

\subsection{Case Studies}
\label{sec:case}
\begin{figure}[hb]
    \centering

    \begin{subfigure}{\linewidth}
        \centering
        \includegraphics[width=1\linewidth]{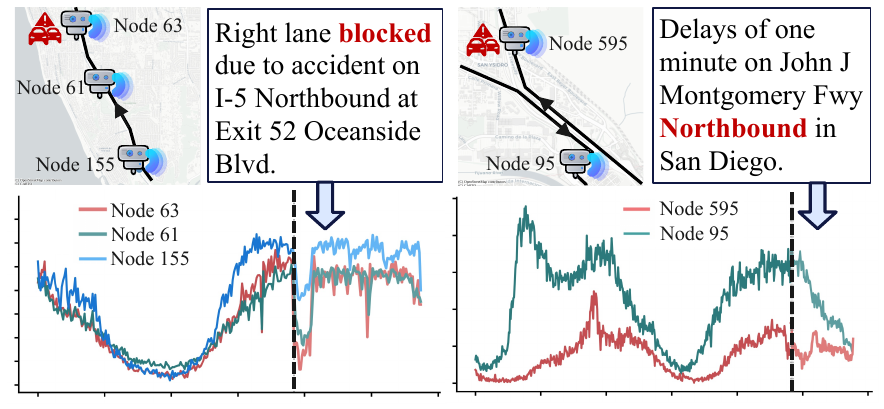}

% across varied directions, intensities of network event evolution
        \caption{
        Network event propagation and varied directions
        }
        \label{fig:case_study_1}
    \end{subfigure}

    \vspace{0.8ex}

    \begin{subfigure}{\linewidth}
        \centering
        \includegraphics[width=1\linewidth]{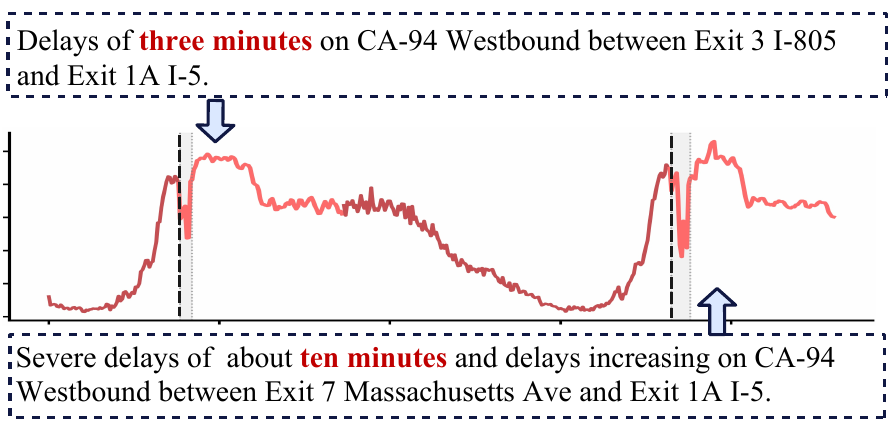}
        \caption{Varied event intensities}
        \label{fig:case_study_2}
    \end{subfigure}

    \caption{Case studies for evaluating the controllability and fidelity of generated simulations. Darker shades denote ground truth (not used as input); lighter shades denote simulations by \method{} (dashed lines).}
    \Description{Two case studies compare event prompts, affected road locations, and simulated traffic curves against ground-truth curves. The first shows impacts propagating to nearby nodes in different directions, and the second shows stronger simulated disruption for a prompt describing longer severe delays.}
    \label{fig:case_study}
\end{figure}

        % Top: Event impact propagates through the network, network traffic drops less as distance increases.
        % Bottom:
To qualitatively understand the controllability and fidelity of generated simulations by \method{}, we present several case studies by contrasting input prompts in Figure~\ref{fig:case_study}:
% To qualitatively demonstrate the effectiveness of \method{}, we present case studies examining three aspects of conditional generation in.
\begin{enumerate}[(a)]
    \item \textbf{Event Propagation and Varied Directions}: Left: Event impacts propagate through the road network, with the largest traffic drop at the directly affected node and weaker but correlated declines at nearby nodes. Node 63, 61, and 155 are progressively farther from the incident center, and \task{} reproduces the corresponding attenuation across these locations.
    Right: A crash in the northbound lanes reduces traffic at northbound sensors while southbound sensors remain unaffected. On the John J Montgomery Fwy in San Diego, Node 595 in the northbound direction shows a sharp drop, whereas southbound Node 95 is largely unchanged. Our model generates these direction-specific patterns from the text and closely matches the ground truth, capturing location-specific spatial cues.

    \item \textbf{Varied intensities}: We compare events with different delay severities on CA-94 Westbound. For a three-minute delay, the generated pattern shows moderate fluctuations, while for a ten-minute severe delay, the model produces larger amplitude variations. This demonstrates that our model effectively interprets intensity-related semantics from the text and reflects them in the generated traffic patterns.
    %% varied types of events??
   \end{enumerate}
These case studies complement the quantitative evidence that \method{} can preserve topology while responding to event semantics.
% (\romannumeral1) 
% (\romannumeral2) 
% (\romannumeral3) 

%% file: tables/overall.tex
\begin{table*}[htbp]
    \centering 
    \caption{Performance comparison on \task{}-6.5M benchmark over different horizons. (N/A indicates that all samples fall below the MRR@5 threshold)}
    \label{tab:overall}
    \setlength{\tabcolsep}{1.5pt}
    \resizebox{\textwidth}{!}{

      \begin{tabular}{cc|cccc|cccc|cccc|cccc}
          \toprule
          \toprule
          & & \multicolumn{4}{c|}{\textbf{SD}} & \multicolumn{4}{c|}{\textbf{GLA}} & \multicolumn{4}{c|}{\textbf{GBA}} & \multicolumn{4}{c}{\textbf{CA}} \\
          \cmidrule(lr){3-6}\cmidrule(lr){7-10}\cmidrule(lr){11-14}\cmidrule(lr){15-18}
          \textbf{Methods} & \textbf{Steps}
          & \textbf{ED}$\downarrow$ & \textbf{C-FID}$\downarrow$ & \textbf{MRR@5}$\uparrow$ & \textbf{JL-MMD}$\downarrow$
          & \textbf{ED}$\downarrow$ & \textbf{C-FID}$\downarrow$ & \textbf{MRR@5}$\uparrow$ & \textbf{JL-MMD}$\downarrow$
          & \textbf{ED}$\downarrow$ & \textbf{C-FID}$\downarrow$ & \textbf{MRR@5}$\uparrow$ & \textbf{JL-MMD}$\downarrow$
          & \textbf{ED}$\downarrow$ & \textbf{C-FID}$\downarrow$ & \textbf{MRR@5}$\uparrow$ & \textbf{JL-MMD}$\downarrow$ \\
          
          \midrule

\multirow{3}{*}{\makecell{\textbf{Diffusion-TS}}}  
                            & 24 &{34.84}&{43.35}&{N/A}&{\underline{1.04}}&{60.16}&{43.31}&{N/A}&{0.92}&{67.28}&{42.23}&{N/A}&{0.91}&{126.60}&{44.93}&{N/A}&{0.96}\\
                            & 48 &{35.11}&{48.50}&{N/A}&{0.91}&{60.72}&{43.65}&{N/A}&{\underline{0.65}}&{67.35}&{46.96}&{N/A}&{0.70}&{126.71}&{45.96}&{N/A}&{\underline{0.78}}\\
                            & 96 &{35.76}&{48.89}&{N/A}&{\underline{0.78}}&{61.22}&{47.46}&{N/A}&{\underline{0.45}}&{67.39}&{49.35}&{N/A}&{0.49}&{127.43}&{45.99}&{N/A}&{\underline{0.56}}\\
          \midrule
          
\multirow{3}{*}{\makecell{\textbf{VerbalTS}}}  
                            & 24 &{34.83}&{44.87}&{N/A}&{1.04}&{60.89}&{42.44}&{N/A}&{0.87}&{67.28}&{41.50}&{N/A}&{\underline{0.89}}&{126.62}&{43.78}&{N/A}&{0.98}\\
                            & 48 &{35.10}&{45.82}&{N/A}&{\underline{0.90}}&{61.72}&{43.82}&{N/A}&{0.66}&{67.34}&{47.16}&{N/A}&{\underline{0.69}}&{126.71}&{45.53}&{N/A}&{0.79}\\
                            & 96 &{35.75}&{50.25}&{N/A}&{0.78}&{62.25}&{45.65}&{N/A}&{0.45}&{67.40}&{48.45}&{N/A}&{\underline{0.47}}&{127.31}&{46.66}&{N/A}&{0.57}\\
          \midrule
          
\multirow{3}{*}{\makecell{\textbf{ChatTraffic}}}  
                            & 24 &{26.17}&{37.59}&{0.43}&{1.06}&{45.58}&{35.04}&{\underline{0.64}}&{\underline{0.86}}&{53.88}&{30.34}&{\underline{0.68}}&{0.90}&{70.83}&{35.78}&{\underline{0.74}}&{\underline{0.88}}\\
                            & 48 &{26.55}&{40.86}&{\underline{0.47}}&{0.93}&{46.42}&{38.51}&{\underline{0.65}}&{0.68}&{54.43}&{31.31}&{\underline{0.71}}&{0.69}&{70.84}&{35.97}&{\underline{0.75}}&{0.85}\\
                            & 96 &{27.45}&{46.15}&{\underline{0.49}}&{0.80}&{47.94}&{38.83}&{\underline{0.66}}&{0.47}&{55.67}&{40.37}&{\underline{0.75}}&{0.48}&{72.83}&{35.99}&{\underline{0.75}}&{0.81}\\
          \midrule

\multirow{3}{*}{\makecell{\textbf{T2S}}} 
                            & 24 &{\underline{21.14}}&{\underline{22.47}}&{\underline{0.45}}&{1.05}&{\underline{38.11}}&{\underline{24.65}}&{0.53}&\underline{0.86}&{\underline{46.15}}&{\underline{29.28}}&{\underline{0.68}}&{0.90}&{\underline{64.57}}&{\underline{22.38}}&{0.73}&{0.96}\\
                            & 48 &{\underline{21.50}}&{\underline{28.32}}&{0.47}&{0.92}&{\underline{38.83}}&{\underline{29.03}}&{0.54}&{0.72}&{\underline{46.21}}&{\underline{29.96}}&{0.70}&{0.76}&{\underline{64.69}}&{\underline{22.52}}&\underline{0.75}&{0.92}\\
                            & 96 &{\underline{22.37}}&{\underline{35.84}}&{0.47}&{0.78}&{\underline{39.10}}&{\underline{29.19}}&{0.58}&{0.53}&{\underline{46.34}}&{\underline{31.65}}&{0.73}&{0.53}&{\underline{65.11}}&{\underline{22.80}}&{0.75}&{0.85}\\
          \midrule
          
\multirow{3}{*}{\makecell{\textbf{\method}}}  
                            & 24 &{\textbf{20.62}}&{\textbf{20.52}}&{\textbf{0.53}}&{\textbf{0.85}}&{\textbf{32.11}}&{\textbf{23.48}}&{\textbf{0.68}}&{\textbf{0.65}}&{\textbf{43.00}}&{\textbf{28.60}}&{\textbf{0.76}}&{\textbf{0.61}}&{\textbf{57.88}}&{\textbf{20.46}}&{\textbf{0.77}}&{\textbf{0.64}}\\
                            & 48 &{\textbf{20.99}}&{\textbf{22.77}}&{\textbf{0.55}}&{\textbf{0.66}}&{\textbf{32.32}}&{\textbf{27.77}}&{\textbf{0.69}}&{\textbf{0.41}}&{\textbf{43.92}}&{\textbf{29.07}}&{\textbf{0.77}}&{\textbf{0.41}}&{\textbf{58.02}}&{\textbf{20.61}}&{\textbf{0.78}}&{\textbf{0.53}}\\
                            & 96 &{\textbf{21.88}}&{\textbf{29.09}}&{\textbf{0.59}}&{\textbf{0.63}}&{\textbf{32.91}}&{\textbf{29.15}}&{\textbf{0.70}}&{\textbf{0.38}}&{\textbf{44.08}}&{\textbf{30.95}}&{\textbf{0.77}}&{\textbf{0.40}}&{\textbf{58.85}}&{\textbf{20.71}}&{\textbf{0.78}}&{\textbf{0.41}}\\
          \bottomrule
          \bottomrule
      \end{tabular}
    }
\end{table*}

%% file: tables/ablation.tex
\begin{table*}[htbp]
    \centering
    \caption{Ablation study of \method~across \task{}-6.5M (over a horizon of 24 steps).}
    \label{tab:ablation}
    \setlength{\tabcolsep}{1.5pt}
    \resizebox{\textwidth}{!}{
    \Large
      \begin{tabular}{c|cccc|cccc|cccc|cccc}
          \toprule
          \toprule
          & \multicolumn{4}{c|}{\textbf{SD}} & \multicolumn{4}{c|}{\textbf{GLA}} & \multicolumn{4}{c|}{\textbf{GBA}} & \multicolumn{4}{c}{\textbf{CA}} \\
          \cmidrule(lr){2-5}\cmidrule(lr){6-9}\cmidrule(lr){10-13}\cmidrule(lr){14-17}
          \textbf{Variants}
          & \textbf{ED}$\downarrow$ & \textbf{C-FID}$\downarrow$ & \textbf{MRR@5}$\uparrow$ & \textbf{JL-MMD}$\downarrow$
          & \textbf{ED}$\downarrow$ & \textbf{C-FID}$\downarrow$ & \textbf{MRR@5}$\uparrow$ & \textbf{JL-MMD}$\downarrow$
          & \textbf{ED}$\downarrow$ & \textbf{C-FID}$\downarrow$ & \textbf{MRR@5}$\uparrow$ & \textbf{JL-MMD}$\downarrow$
          & \textbf{ED}$\downarrow$ & \textbf{C-FID}$\downarrow$ & \textbf{MRR@5}$\uparrow$ & \textbf{JL-MMD}$\downarrow$ \\
          
          \midrule

          \textbf{W/o $\mathbf{M}^{(e)}$} &{25.32}&{33.33}&{0.47}&{0.99}&{41.30}&{30.34}&{0.54}&{0.88}&{53.55}&{31.36}&{0.68}&{0.85}&{65.10}&{34.36}&{0.64}&{0.86}\\
          \textbf{W/o $\mathbf{M}^{(t,s)}$} &{25.06}&{31.38}&{0.49}&{0.99}&{39.25}&{29.54}&{0.53}&{0.87}&{51.35}&{30.55}&{0.71}&{0.83}&{64.55}&{34.56}&{0.65}&{0.85}\\
          \textbf{W/o $\mathbf{E}_d$} &{24.63}&{30.80}&{0.50}&{0.96}&{35.57}&{27.60}&{0.55}&{0.79}&{51.14}&{30.99}&{0.70}&{\underline{0.82}}&{63.64}&{28.24}&{0.67}&{\underline{0.82}}\\
          \textbf{W/o $\mathbf{E}_d^{(w)}$} &{24.52}&{28.67}&{0.50}&{\underline{0.95}}&{34.21}&{28.33}&{\underline{0.56}}&{0.79}&{49.55}&{29.86}&{\underline{0.73}}&{0.82}&{60.81}&{28.21}&{0.67}&{0.83}\\
          \textbf{W/o $\mathbf{E}_d^{(i)}$} &{24.50}&{28.66}&{\underline{0.51}}&{0.95}&{34.20}&{27.31}&{0.56}&{\underline{0.78}}&{49.55}&{29.84}&{0.73}&{0.83}&{60.93}&{26.47}&{\underline{0.69}}&{0.84}\\
          \textbf{W/o Graph U-Net} &{\underline{23.74}}&{\underline{26.64}}&{0.51}&{1.01}&{\underline{34.19}}&{\underline{27.03}}&{0.55}&{0.90}&{\underline{49.53}}&{\underline{29.74}}&{0.73}&{0.94}&{\underline{60.12}}&{\underline{24.20}}&{0.68}&{1.04}\\
          \midrule
          \textbf{\method} &{\textbf{20.62}}&{\textbf{20.52}}&{\textbf{0.53}}&{\textbf{0.85}}&{\textbf{32.11}}&{\textbf{23.48}}&{\textbf{0.58}}&{\textbf{0.65}}&{\textbf{43.00}}&{\textbf{28.60}}&{\textbf{0.76}}&{\textbf{0.61}}&{\textbf{57.88}}&{\textbf{20.46}}&{\textbf{0.77}}&{\textbf{0.64}}\\
          \bottomrule
          \bottomrule
      \end{tabular}
    }
\end{table*}

%% file: tables/cttp.tex
\begin{table}[ht]
    \centering
    \caption{Condition-following evaluation using CTTP retrieval accuracy (\%, higher the better). GT denotes the score computed on ground-truth event--traffic pairs.}
    \label{tab:cttp}
    \setlength{\tabcolsep}{2.2pt}
    \resizebox{\columnwidth}{!}{
    \begin{tabular}{l|ccccc|c}
        \toprule
        \toprule
        \textbf{Dataset} & \textbf{Diffusion-TS} & \textbf{VerbalTS} & \textbf{ChatTraffic} & \textbf{T2S} & \textbf{\method} & \textbf{GT} \\
        \midrule
        GBA & \underline{35.20} & 29.85 & 31.24 & 30.54 & \textbf{37.93} & 54.51 \\
        GLA & 31.13 & 32.52 & \underline{33.65} & 30.52 & \textbf{40.86} & 59.07 \\
        SD  & 24.37 & 31.27 & \underline{34.73} & 31.23 & \textbf{44.72} & 60.72 \\
        \bottomrule
        \bottomrule
    \end{tabular}
    }
\end{table}

%% file: tables/generalization.tex
\begin{table}[htbp]
    \centering
    \caption{Generalization study across years and networks (24 steps). FT: Finetune, FS: Fewshot. Y: Year (temporal transfer across year), N: Number of nodes (cross-network transfer). MRR@5 is omitted as all models fall below the threshold.}
    \label{tab:generalization}
    \resizebox{\columnwidth}{!}{
    \begin{tabular}{cc|l|ccc}
        \toprule
        \toprule
        \textbf{Setting} & \textbf{Transfer} & \textbf{Method} & \textbf{ED}$\downarrow$ & \textbf{C-FID}$\downarrow$ & \textbf{JL-MMD}$\downarrow$ \\
        \midrule
        \multirow{5}{*}{\textbf{FT}}
        & \multirow{5}{*}{\shortstack{GLA \\ Y2017 \\ $\downarrow$ \\ GLA \\ Y2018}}
        & Diffusion-TS &{164.82}&{512.37}&{0.94}\\
        & & VerbalTS &{101.13}&{185.32}&{0.94}\\
        & & ChatTraffic &{155.14}&{405.56}&{\underline{0.93}}\\
        & & T2S &{\underline{61.02}}&{\underline{73.12}}&{0.94}\\
        & & \method &{\textbf{55.93}}&{\textbf{60.45}}&{\textbf{0.89}}\\
        \cmidrule{1-6}
        \multirow{5}{*}{\textbf{FS}}
        & \multirow{5}{*}{\shortstack{GLA \\ Y2017 \\ $\downarrow$ \\ GLA \\ Y2018}}
        & Diffusion-TS & {270.60} & {1751.29} & {0.95} \\
        & & VerbalTS & {101.22} & {162.33} & \underline{0.93} \\
        & & ChatTraffic & {159.14} & {512.95} & \underline{0.93} \\
        & & T2S & \underline{60.46} & \underline{71.72} & \underline{0.93} \\
        & & \method & \textbf{57.80} & \textbf{63.04} & \textbf{0.88} \\
                
        \midrule
        \midrule

        \multirow{10}{*}{\textbf{FT}}
        & \multirow{5}{*}{\shortstack{SD \\ N705 \\ $\downarrow$ \\ GLA \\ N1820}}
        & Diffusion-TS &{196.72}&{922.53}&{0.89}\\
        & & VerbalTS &{100.34}&{\underline{73.31}}&{0.89}\\
        & & ChatTraffic &{114.44}&{152.58}&{\underline{0.88}}\\
        & & T2S &{\underline{73.24}}&{75.63}&{0.91}\\
        & & \method &{\textbf{65.27}}&{\textbf{70.47}}&{\textbf{0.78}}\\
        \cmidrule{2-6}
        & \multirow{5}{*}{\shortstack{GBA \\ N2294 \\ $\downarrow$ \\ GLA \\ N1820}}
        & Diffusion-TS &{182.95}&{391.47}&{\underline{0.89}}\\
        & & VerbalTS &{100.29}&{79.00}&{0.89}\\
        & & ChatTraffic &{134.33}&{225.76}&{0.89}\\
        & & T2S &{\underline{73.71}}&{\underline{73.57}}&{0.91}\\
        & & \method &{\textbf{64.39}}&{\textbf{70.50}}&{\textbf{0.77}}\\
        
        \midrule

        \multirow{10}{*}{\textbf{FS}}
        & \multirow{5}{*}{\shortstack{SD \\ N705 \\ $\downarrow$ \\ GLA \\ N1820}}
        & Diffusion-TS &{292.41}&{1032.85}&{0.90}\\
        & & VerbalTS &{100.34}&{\underline{80.86}}&{0.90}\\
        & & ChatTraffic &{125.08}&{187.55}&{\underline{0.88}}\\
        & & T2S &{\underline{76.67}}&{84.97}&{0.91}\\
        & & \method &{\textbf{70.75}}&{\textbf{77.83}}&{\textbf{0.76}}\\
        \cmidrule{2-6}
        & \multirow{5}{*}{\shortstack{GBA \\ N2294 \\ $\downarrow$ \\ GLA \\ N1820}}
        & Diffusion-TS &{223.36}&{840.04}&{0.90}\\
        & & VerbalTS &{100.28}&{75.87}&{0.89}\\
        & & ChatTraffic &{150.12}&{240.25}&{\underline{0.88}}\\
        & & T2S &{\underline{68.98}}&{\underline{75.13}}&{0.90}\\
        & & \method &{\textbf{65.12}}&{\textbf{74.66}}&{\textbf{0.77}}\\

        \bottomrule
        \bottomrule
    \end{tabular}
    }
\end{table}

%% file: tables/psme.tex
\begin{table}[ht]
    \centering
    \caption{Cross-domain validation on PSME (96 steps)}
    \label{tab:psme}
    \setlength{\tabcolsep}{2.2pt}
    \resizebox{\columnwidth}{!}{
    \begin{tabular}{l|ccccc}
        \toprule
        \toprule
        \textbf{Metric} & \textbf{Diffusion-TS} & \textbf{VerbalTS} & \textbf{ChatTraffic} & \textbf{T2S} & \textbf{\method} \\
        \midrule
        ED$\downarrow$ & \underline{8.19} & 8.21 & 9.39 & 8.96 & \textbf{5.14} \\
        C-FID$\downarrow$ & 13.19 & 13.34 & 13.17 & \underline{11.35} & \textbf{8.59} \\
        JL-MMD$\downarrow$ & 1.30 & 1.30 & 1.31 & \underline{1.18} & \textbf{0.97} \\
        \bottomrule
        \bottomrule
    \end{tabular}
    }
\end{table}

%% file: tables/efficiency_tables.tex
\begin{table}[htbp]
    \centering
    \caption{Training efficiency analysis on the generative diffusion models (in ms/batch, 96 steps).}
    \label{tab:efficiency}
    \resizebox{\columnwidth}{!}{
    \begin{tabular}{l|cccc}
        \toprule
        \toprule
        \textbf{Model} & \textbf{SD} & \textbf{GLA} & \textbf{GBA} & \textbf{CA} \\
        \midrule
        \textbf{Diffusion-TS} & 95.41{\scriptsize$\pm$196.26} & 234.57{\scriptsize$\pm$12.44} & 337.88{\scriptsize$\pm$20.44} & 2045.95{\scriptsize$\pm$44.18} \\
        \textbf{VerbalTS} & 60.70{\scriptsize$\pm$14.22} & 166.97{\scriptsize$\pm$9.61} & 222.27{\scriptsize$\pm$2.77} & 1750.51{\scriptsize$\pm$16.95} \\
        \textbf{ChatTraffic} & 88.75{\scriptsize$\pm$5.04} & 351.24{\scriptsize$\pm$3.09} & 501.58{\scriptsize$\pm$5.28} & 5358.36{\scriptsize$\pm$57.03} \\
        \textbf{T2S} & 97.74{\scriptsize$\pm$9.41} & 617.45{\scriptsize$\pm$3.13} & 957.62{\scriptsize$\pm$3.55} & 13569.12{\scriptsize$\pm$89.38} \\
        \textbf{\method} & \textbf{51.44{\scriptsize$\pm$23.93}} & \textbf{142.68{\scriptsize$\pm$12.03}} & \textbf{191.90{\scriptsize$\pm$11.26}} & \textbf{1744.99{\scriptsize$\pm$48.43}} \\
        \bottomrule
        \bottomrule
    \end{tabular}
    }
\end{table}

%% file: content/6_Related_Work.tex
\section{Related Work}

\subsection{Time Series Generation}

Early Deep generative models for time series approaches , such as VRAE~\cite{fabius2015variationalrecurrentautoencoders}, employed VAEs with RNNs to learn latent representations of temporal dynamics, enabling the generation of sequences from prior distributions.
Generative adversarial networks (GANs) subsequently demonstrated strong performance by training a generator and discriminator in an adversarial fashion to produce more diverse temporal patterns. 
Notable examples include TimeGAN~\cite{NEURIPS2019_c9efe5f2}, which combines adversarial training with a supervised loss on stepwise embeddings to jointly preserve temporal dynamics and distributional fidelity.
More recently, DDPMs~\cite{ge2025t2s} have emerged as the dominant paradigm for generative modeling, owing to their stable training and superior sample quality compared to GANs. In the time series domain, TimeGrad~\cite{Rasul2021AutoregressiveDD} incorporates the denoising process into an autoregressive framework for probabilistic forecasting, while CSDI~\cite{10.5555/3540261.3542161} demonstrates the effectiveness of score-based diffusion for time series imputation by conditioning on observed values.
SSD-TS~\cite{10.1145/3711896.3737135} uses Mamba-based linear state space models to make diffusion time-series imputation faster and better at capturing temporal dependencies.
Beyond single-modality generation, the success of text-conditioned synthesis in vision and audio domains-such as text-to-image~\cite{rombach2022ldm}, text-to-video~\cite{singer2022make}, and text-to-audio~\cite{huang2023make,liu2023audioldm} has inspired cross-modal approaches for time series.
These methods leverage natural language descriptions to guide temporal data synthesis, enabling more flexible and semantically controllable generation. Recent works such as VerbalTS~\cite{gu2025verbalts} and T2S~\cite{ge2025t2s} adapt diffusion-based architectures with text encoders to generate time series from textual descriptions.
However, these methods typically condition only on an event’s text and thus ignore the event’s multi-structural information, a key challenge in network event evolution.

\subsection{Graph Neural Networks}

Graph neural networks (GNNs) have become the standard tool for learning on graph-structured data, as they explicitly encode relational inductive bias and propagate information along observed connectivity. Spectral approaches such as ChebNet\cite{NEURIPS2022_2f9b3ee2} approximate graph convolutions via Chebyshev polynomials, and GCN\cite{kipf2017semisupervised} further simplifies this formulation into a first-order approximation, enabling efficient neighborhood aggregation on large graphs. Spatial methods including GraphSAGE\cite{10.5555/3294771.3294869} and GAT\cite{2018graphat} generalize message passing through sampling-based aggregation and attention-weighted neighbors, respectively.
To process graphs at multiple scales, hierarchical pooling operators have been proposed to capture both local interactions and broader structural dependencies. DiffPool~\cite{10.5555/3327345.3327389} learns soft cluster assignments to coarsen graphs in an end-to-end fashion, while gPool~\cite{10.1145/3308558.3313395} selects top-$k$ nodes based on learnable projection scores, offering a simple yet effective hard-selection alternative. Graph U-Net~\cite{gao2019graph}, inspired by the U-Net architecture~\cite{10.1007/978-3-319-24574-4_28} in image segmentation, pairs gPool with an unpooling operator and skip connections to form an encoder-decoder structure that preserves fine-grained node information across hierarchical levels. This property is particularly useful for network evolution simulation, where event impacts may first appear locally but then propagate through multi-hop connectivity.
We incorporate pooling–unpooling operations into a diffusion framework to enable conditional modulation and to preserve network topology throughout the denoising process.

\subsection{Network Traffic Simulation}

Network traffic simulation has traditionally relied on mechanistic traffic-flow models and calibrated simulators. Macroscopic formulations, such as multi-class extensions of the LWR model~\cite{wong2002multi}, describe how traffic density and flow evolve through conservation laws, while widely used microscopic simulators such as SUMO~\cite{krajzewicz2012recent} and Vissim~\cite{vissim2022ptv} provide detailed vehicle-level dynamics. These systems are valuable for controlled analysis, but they usually require hand-crafted assumptions, expert calibration, and scenario-specific configuration, which limits their scalability for large numbers of diverse event conditions.

Learning-based traffic modeling has largely focused on forecasting from observed histories. Representative spatio-temporal predictors use graph diffusion~\cite{li2018diffusion}, attention over road networks~\cite{Zheng_Fan_Wang_Qi_2020}, or incident-aware graph convolutions~\cite{xie2020deep} to estimate future traffic states. Although effective for prediction, they are not designed to generate event-conditioned network evolution from natural-language descriptions, nor do they model a distribution of possible future responses under sparse or partially structured event inputs. Generative traffic models move closer to simulation, including off-deployment traffic estimation~\cite{zhang2019trafficgan}, conditional urban traffic estimation~\cite{zhang2020curb}, and knowledge-enhanced diffusion for urban flow generation~\cite{zhou2023towards}. Recent promptable or language-driven simulators further improve controllability for traffic scenarios and trajectories~\cite{tan2024promptable,NEURIPS2023_d95cb79a,xia2024languagedriven}. However, most of these methods target flow estimation, vehicle-level scenarios, or trajectory generation, rather than simulating how event impacts propagate over a fixed road-network topology. 
% In contrast, \method{} aligns event descriptions with network traffic snapshots and uses topology-aware diffusion to generate multi-step network event evolution.

%% file: content/7_Conclusion.tex
\section{Conclusion}

In this paper, we presented \task{}, a diffusion-based framework for simulating network dynamics under event conditions. By coupling a structure-guided masked autoencoder with a Graph U-Net denoiser, our approach bridged multi-structural event conditioning and topology-preserving generation---two aspects that prior diffusion models largely neglected. We further introduced JL-MMD and constructed a multimodal benchmark with over 6.5 million event--traffic pairs. One limitation is that the current model is primarily designed upon transductive GNNs. Network expansions or node removals are implicitly handled by updating adjacency matrices and masking affected features, while an adapter can better align new nodes with the learnt node space. An inductive variant could be developed by replacing the GCN backbone with GraphSAGE-style aggregation while retaining pooling and unpooling. Another direction is to extend the framework to multi-round generation, enabling the simulation of cascading and compound events.

% An inductive version could replace the GCN backbone with GraphSAGE-style aggregation while retaining pooling and unpooling. In the future, extending the framework to multi-round generation for simulating cascading compound events is another promising direction. 

%% file: content/9_arxiv.tex
\newcommand{\arxivjlmmdanalysisextra}{
\subsection{Additional Details on Topology Encoding}
\label{app:graph-embedding}

Given a sensor network $G=(V,\mathbf{A})$ with $N$ nodes and observations $\mathbf{X}\in\mathbb{R}^{T\times N}$, JL-MMD first maps each network trajectory to a graph-aware embedding. For node $v_i\in V$, the embedding aggregates temporal features from its adjacency neighbors:
\begin{equation}
    \mathbf{h}_i=\sum_{j\in\mathcal{N}(i)\cup\{i\}}\mathbf{A}_{ij}\cdot f(\mathbf{X}_{:,j}),
\end{equation}
where $\mathcal{N}(i)$ denotes the neighbors of node $i$, and $f(\cdot)$ extracts temporal features. A graph-level representation is obtained by aggregating node embeddings:
\begin{equation}
    \mathbf{h}_G=\mathrm{Agg}(\{\mathbf{h}_i\}_{i=1}^{N})\in\mathbb{R}^{d}.
\end{equation}
This construction makes the metric sensitive to topology: two trajectories with the same marginal node values can still receive different graph embeddings if their connectivity or spatial assignment differs.

\noindent\textbf{RBF kernel extension.}
The main experiments use the linear kernel for scalability. For completeness, the same JL projection argument also applies to the RBF kernel $\kappa_{\mathrm{rbf}}(\mathbf{x},\mathbf{y})=\exp(-\gamma\|\mathbf{x}-\mathbf{y}\|_2^2)$. Since $g(t)=\exp(-\gamma t)$ is $\gamma$-Lipschitz, the projected MMD error satisfies
\begin{equation}
    \left|\mathrm{MMD}^{2}_{\mathrm{rbf}}(\boldsymbol{\Phi}\mathcal{P},\boldsymbol{\Phi}\mathcal{Q})-
    \mathrm{MMD}^{2}_{\mathrm{rbf}}(\mathcal{P},\mathcal{Q})\right|
    \leq 4\gamma\varepsilon D_{\max}^{2}+\delta_n .
\end{equation}
RBF is characteristic but requires pairwise kernel evaluation, while the linear kernel gives the practical $O(nd)$ estimator used for large-scale experiments.
}

\newcommand{\arxivsyntheticvalidationextra}{
For reproducibility, signals on each node is generated by
\begin{equation}
    \mathbf{X}_{t,i}=\sin(\omega t)+0.1t+\sum_{j\in\mathcal{N}(i)}0.3\mathbf{A}_{ij}\mathbf{X}_{t,j}+\epsilon_{t,i},
\end{equation}
where $\omega=4\pi/T$ controls periodicity and $\epsilon_{t,i}\sim\mathcal{N}(0,0.1)$ is Gaussian noise. The graph diffusion term introduces spatial correlations tied to $\mathbf{A}$, allowing node values, topology, or their assignment to be perturbed independently.

In addition to ED, C-FID, and MRR@5, we report the following point-wise accuracy metrics, which are included only in this synthetic validation: ($\epsilon=10^{-8}$ is set to avoid zero division)
\begin{align}
    \mathrm{MAE}(\mathbf{X},\hat{\mathbf{X}})&=\frac{1}{TN}\sum_{t=1}^{T}\sum_{n=1}^{N}|\mathbf{X}_{t,n}-\hat{\mathbf{X}}_{t,n}|,\\
    \mathrm{RMSE}(\mathbf{X},\hat{\mathbf{X}})&=\sqrt{\frac{1}{TN}\sum_{t=1}^{T}\sum_{n=1}^{N}(\mathbf{X}_{t,n}-\hat{\mathbf{X}}_{t,n})^2},\\
    \mathrm{MAPE}(\mathbf{X},\hat{\mathbf{X}})&=\frac{100}{TN}\sum_{t=1}^{T}\sum_{n=1}^{N}\frac{|\mathbf{X}_{t,n}-\hat{\mathbf{X}}_{t,n}|}{|\mathbf{X}_{t,n}|+\epsilon},\\
    \mathrm{WAPE}(\mathbf{X},\hat{\mathbf{X}})&=\frac{\sum_{t=1}^{T}\sum_{n=1}^{N}|\mathbf{X}_{t,n}-\hat{\mathbf{X}}_{t,n}|}{\sum_{t=1}^{T}\sum_{n=1}^{N}|\mathbf{X}_{t,n}|}.
\end{align}
}

\input{content/8_CameraRready}

\section{Additional Details on Experimental Setup}
\label{app:experimental-setup}

\subsection{Metric Definitions}
\label{app:metrics}

% This section provides additional formulations for metrics used in the experiments. Let $\mathbf{X}\in\mathbb{R}^{T\times N}$ denote the ground-truth traffic sequence and $\hat{\mathbf{X}}\in\mathbb{R}^{T\times N}$ denote the generated sequence.

\noindent\textbf{ED} measures the Euclidean distance between $\mathbf{X}$ and $\hat{\mathbf{X}}$ in feature space:
\begin{equation}
    \mathrm{ED}(\mathbf{X},\hat{\mathbf{X}})=\|\mathbf{X}-\hat{\mathbf{X}}\|_F=
    \sqrt{\sum_{t=1}^{T}\sum_{n=1}^{N}(\mathbf{X}_{t,n}-\hat{\mathbf{X}}_{t,n})^2}.
\end{equation}

\noindent\textbf{C-FID} computes Fr\'echet distance in the latent space of a pre-trained time series encoder, for which we use TS2Vec~\cite{ts2vec}.
\begin{equation}
    \mathrm{C\text{-}FID}=\|\boldsymbol{\mu}_r-\boldsymbol{\mu}_g\|_2^2+
    \mathrm{Tr}\left(\boldsymbol{\Sigma}_r+\boldsymbol{\Sigma}_g-2(\boldsymbol{\Sigma}_r\boldsymbol{\Sigma}_g)^{1/2}\right),
\end{equation}
where $(\boldsymbol{\mu}_r,\boldsymbol{\Sigma}_r)$ and $(\boldsymbol{\mu}_g,\boldsymbol{\Sigma}_g)$ are the empirical mean and covariance of real and generated representations.

\noindent\textbf{MRR@5} measures how reliably the model reproduces the ground truth across 5 independent runs with different random seeds. For each query event $i$, we generate five candidate sequences and compute their cosine similarity to the ground truth in TS2Vec latent space. We sort the five candidates by descending similarity, so $\mathrm{rank}_i$ denotes the position of the first candidate whose similarity exceeds threshold $\tau=0.5$. The metric averages the reciprocal of this index of rank over all queries:
\begin{equation}
    \mathrm{MRR@5}=\frac{1}{|\mathcal{Q}|}\sum_{i=1}^{|\mathcal{Q}|}\frac{1}{\mathrm{rank}_i}.
\end{equation}
A higher value indicates that a faithful generation appears among the top positions of the 5 runs, reflecting both generation quality and run-to-run stability. If no run reaches the threshold, the reciprocal rank of this query is set to 0.

\subsection{Hyperparameter Settings}
\label{app:hyperparameters}

Supplementary hyperparameter configurations in Structure-guided MAE pretraining and diffusion model training are presented in Table~\ref{tab:hyperparams}; parameters described in the main text are omitted for brevity.

\noindent\textbf{Optimization Parameters.} Weight decay is set to 0.0 for the diffusion stage to avoid over-regularizing the latent denoising space. We use a OneCycleLR scheduler with maximum learning rate $10^{-4}$ to improve convergence stability, and fix the random seed to 42 across both stages for reproducibility.

\noindent\textbf{Architecture Parameters.} In pretraining, the block hidden size, residual depth, and compression factor control the capacity and bottleneck of Structure-guided MAE. In the diffusion stage, GCN hidden channels define the feature dimension for spatial message passing, while multi-head attention models temporal dependencies inside the Graph Transformer Block.

\noindent\textbf{VAE and Conditioning Parameters.} The KL weight balances reconstruction fidelity and smooth latent regularization. We use BERT-base-uncased as the text encoder, producing 768-dimensional event embeddings that are injected into the denoiser through AdaLN at the bottleneck of Graph U-Net.

\begin{center}
\centering
\captionof{table}{Supplementary hyperparameter configurations}
\label{tab:hyperparams}
\resizebox{\columnwidth}{!}{%
\begin{tabular}{@{}llcc@{}}
\toprule
\toprule
\textbf{Component} & \textbf{Hyperparameter} & \textbf{Pretrain Stage} & \textbf{Diffusion Stage} \\
\midrule
\multirow{4}{*}{\textbf{Optimization}}
& Weight Decay & -- & 0.0 \\
& LR Scheduler & -- & OneCycleLR \\
& Max Learning Rate & -- & 1e-4 \\
& Random Seed & 42 & 42 \\
\addlinespace
\multirow{6}{*}{\textbf{Architecture}}
& Block Hidden Size & 128 & -- \\
& Residual Layers & 2 & -- \\
& Residual Hidden Size & 32 & -- \\
& Compression Factor & 4 & -- \\
& GCN Hidden Channels & -- & 64 \\
& Attention Heads & -- & 4 \\
\addlinespace
\multirow{2}{*}{\textbf{VAE}}
& KL Weight ($w$) & 0.25 & -- \\
& Pretraining Epochs & 200 & -- \\
\addlinespace
\multirow{2}{*}{\textbf{Conditioning}}
& Text Encoder & -- & BERT-base-uncased \\
& Text Embedding Dim & -- & 768 \\
\bottomrule
\bottomrule
\end{tabular}%
}
\end{center}

\section{\task{}-6.5M Benchmark Construction}
\label{app:benchmark-construction}

\subsection{Data Alignment Pipeline}
\label{app:data-preprocessing}

Algorithm~\ref{alg:data_alignment} summarizes the data alignment pipeline between network sensors and events, including traffic incidents and weather events.

\begin{algorithm}[ht]
    \caption{Data Alignment Pipeline}
    \label{alg:data_alignment}
    \footnotesize
    % \vspace{-0.5em}
    \begin{algorithmic}[1]
    \setlength{\itemsep}{0pt}
    \setlength{\parskip}{0pt}
    \REQUIRE Traffic sensors $\mathcal{S}=\{s_i\}_{i=1}^{N_s}$ with coordinates $(lat_i,lng_i)$ and direction $dir_i$
    \REQUIRE Incidents $\mathcal{I}=\{e_j\}_{j=1}^{N_e}$ with coordinates, street $street_j$, and time interval $[t^e_{start},t^e_{end}]$
    \REQUIRE Weather events $\mathcal{W}=\{w_k\}_{k=1}^{N_w}$ with coordinates $(lat_k,lng_k)$, time interval $[t^w_{start},t^w_{end}]$, and type $type^w_k$
    \REQUIRE Buffer distance $\tau=100$m
    \ENSURE Aligned event-sensor pairs

    \STATE \textbf{// Stage 1: Incident Alignment}
    \STATE Project $\mathcal{S}$ and $\mathcal{I}$ to EPSG:3857
    \FORALL{sensor $s_i\in\mathcal{S}$}
        \STATE $B_i\leftarrow\textsc{Buffer}(s_i,\tau)$
        \STATE $\mathcal{I}_i\leftarrow\{e_j\in\mathcal{I}:e_j\cap B_i\neq\emptyset\}$
        \FORALL{$e_j\in\mathcal{I}_i$}
            \STATE $dir_e\leftarrow\textsc{ParseDirection}(street_j)$
            \IF{$dir_i=dir_e$}
                \STATE Align $[t^e_{start},t^e_{end}]$ to sensor time intervals
                \STATE Store aligned incident-sensor pair
            \ENDIF
        \ENDFOR
    \ENDFOR

    \STATE \textbf{// Stage 2: Weather Alignment}
    \STATE $\mathcal{P}\leftarrow\{(lat_k,lng_k)\}_{k=1}^{N_w}$
    \STATE $\mathcal{V}\leftarrow\textsc{VoronoiTessellation}(\mathcal{P})$
    \FORALL{sensor $s_i\in\mathcal{S}$}
        \STATE $v_i\leftarrow\textsc{FindEnclosingCell}(s_i,\mathcal{V})$
        \STATE $w_{v_i}\leftarrow$ weather station associated with $v_i$
        \FORALL{weather event at $w_{v_i}$}
            \STATE Align $[t^w_{start},t^w_{end}]$ to sensor time intervals
            \STATE Store aligned weather-sensor pair
        \ENDFOR
    \ENDFOR
    \RETURN Aligned event-sensor pairs
    \end{algorithmic}
    \vspace{-0.6em}
\end{algorithm}

\subsection{Visualization Weather Alignment}
 
Voronoi tessellation is used to associate network sensors with weather stations without imposing any meteorological assumptions. Figure~\ref{fig:weather_alignment} illustrates the alignment results across the four networks, in supplement to the procedure described in Section~\ref{sec:data_construction}.
\begin{figure*}[!b]
    \centering
    \setlength{\tabcolsep}{2pt}
    \renewcommand{\arraystretch}{0.9}
    \scriptsize
    \begin{tabular}{cccc}
    \includegraphics[width=0.2115\textwidth]{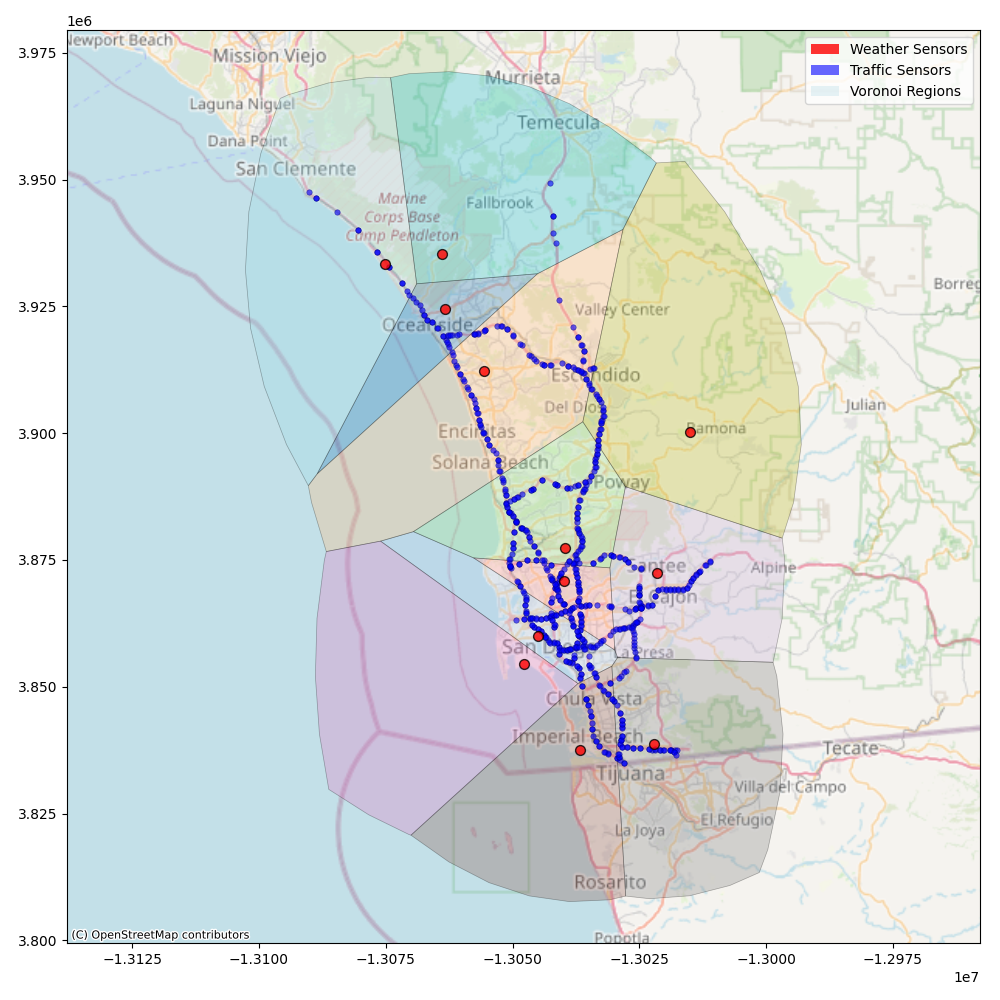} &
    \includegraphics[width=0.2115\textwidth]{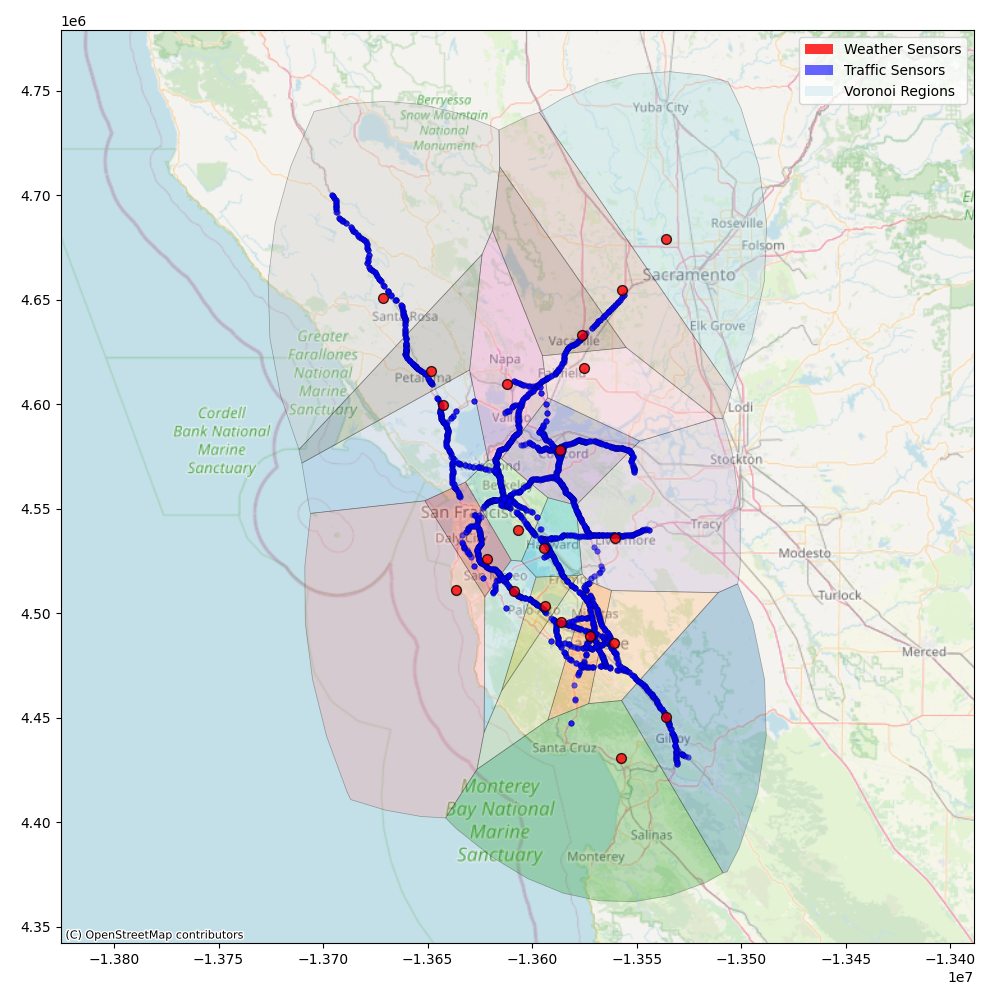} &
    \includegraphics[width=0.2115\textwidth]{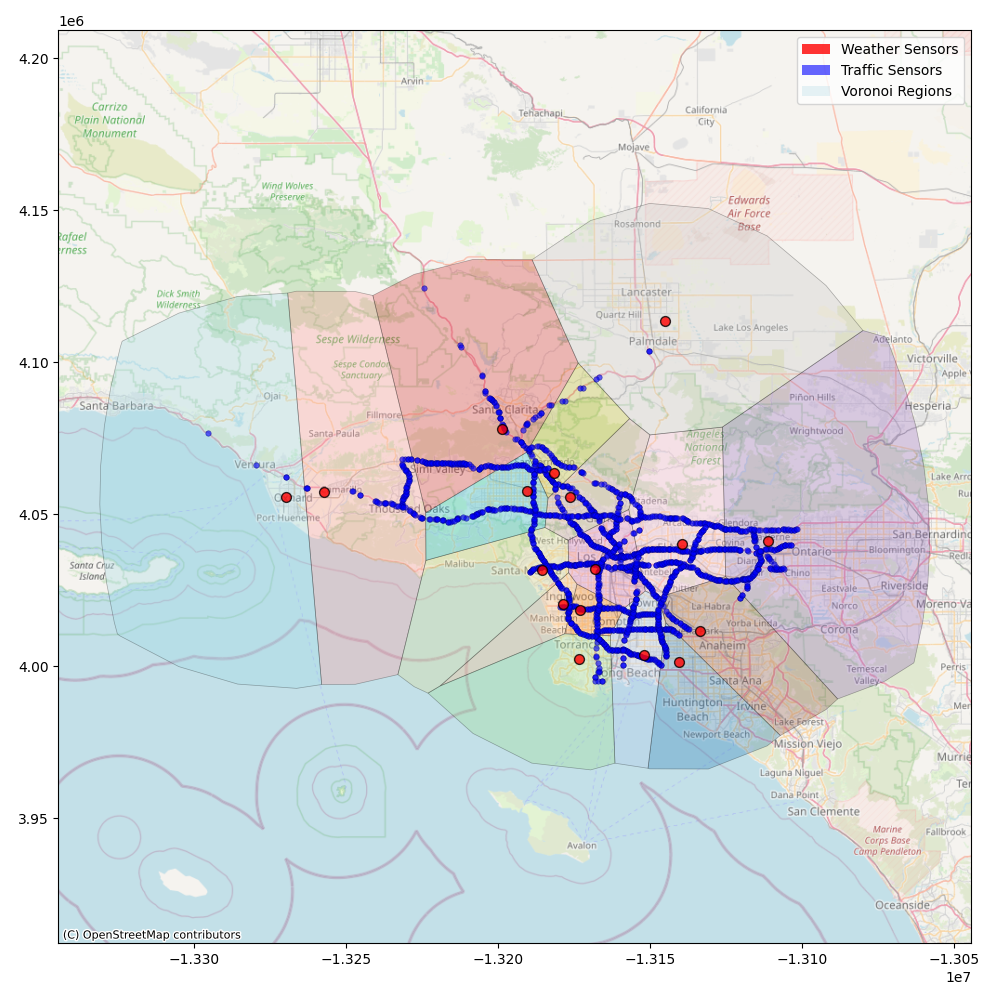} &
    \includegraphics[width=0.2115\textwidth]{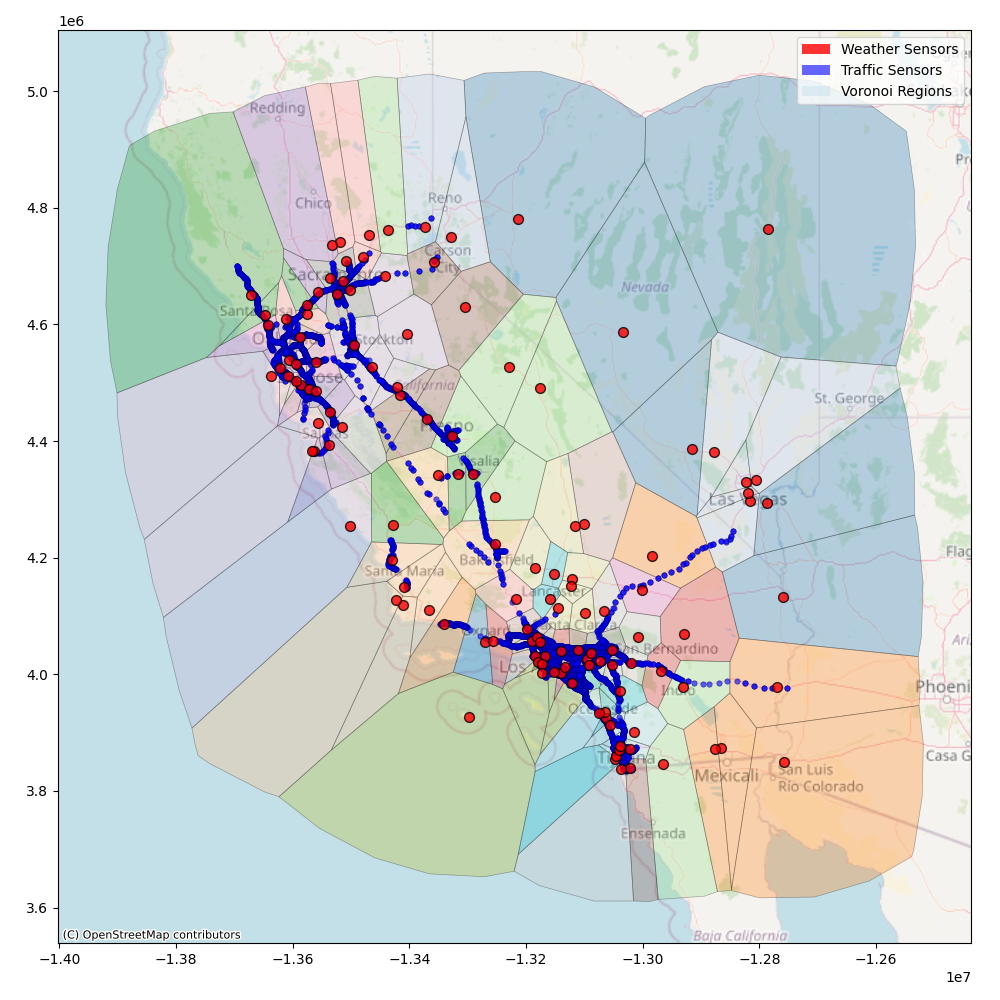} \\
    (a) SD & (b) GBA & (c) GLA & (d) CA
    \end{tabular}
    \vspace{-0.4em}
    \caption{Visualizing weather alignment: Network sensors (blue) are assigned to weather stations (red) based on the Voronoi diagrams}
    \Description{Four maps show weather station assignment regions for SD, GBA, GLA, and CA. Blue traffic sensor points are grouped by colored Voronoi cells around red weather stations, indicating which station supplies weather events for each sensor.}
    \label{fig:weather_alignment}
    \vspace{-0.6em}
    \end{figure*}

\subsection{Weather Event Description Generation}
\label{app:weather-description}
LSTW~\cite{lstw_pattern_discovery} provides textual descriptions for traffic incidents but not for weather events. We leverages LLMs to generate location-aware weather descriptions in three steps:

\noindent\textbf{Step 1: Map Imagery Acquisition.} For each weather station, we use its coordinates $(lat,lng)$ to retrieve a satellite or street map image from Google Maps Static API. We use zoom level 15 to capture local geographic context of roughly 1 km $\times$ 1 km.

\noindent\textbf{Step 2: Geographic Location Extraction.} We formulate geographic location extraction as a visual question answering (VQA) task. We use \texttt{Gemini-3-Pro} with a VQA prompt that contains both an image input and text instructions, asking the model to identify road names, intersections, landmarks, and location-specific features from each map image (as illustrated below).

\noindent\textbf{Step 3: Description Concatenation.} We further concatenate the extracted geographic description with the weather event type:
\begin{quote}
\textit{``\{weather\_type\} weather event occurred at \{location\_description\}.''}
\end{quote}
For example: \textit{``Rain weather event occurred at the intersection of Mission Bay Drive and Ingraham Street, near SeaWorld San Diego.''}

\begin{tcolorbox}[
    colback=aclback,
    colframe=aclbluer,
    coltitle=white,
    title=\textbf{Prompt for Geographic Location Extraction},
    fonttitle=\bfseries,
    fontupper=\footnotesize,
    boxrule=0.6pt,
    arc=1mm,
    left=4pt,
    right=4pt,
    top=3pt,
    bottom=3pt,
    before skip=4pt,
    after skip=6pt
  ]
  \textbf{Image Input:}
  \begin{center}
    \includegraphics[width=0.50\linewidth]{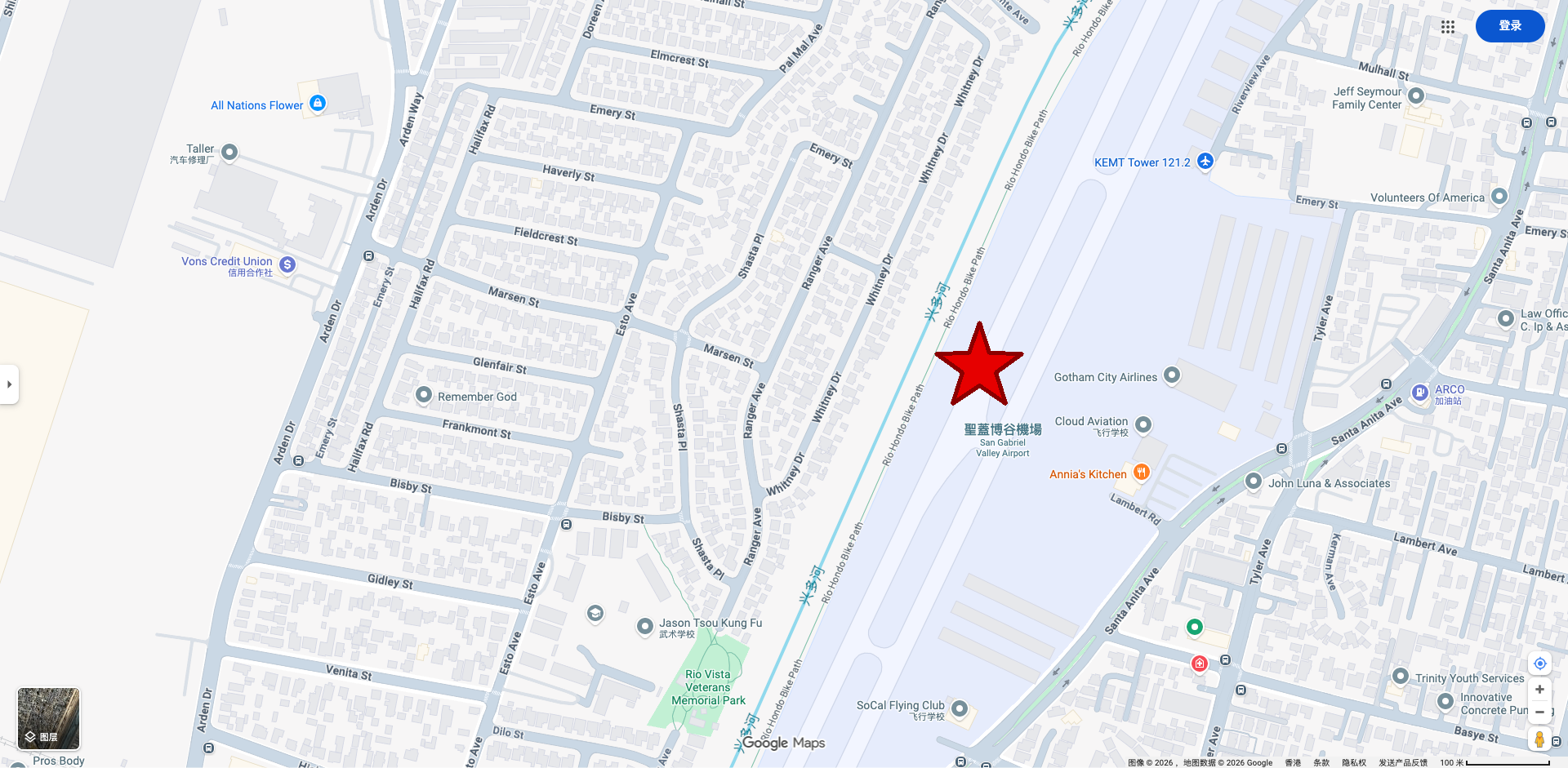}
  \end{center}

  \textbf{Role:} You are a geographic location analyst who specializes in identifying and describing landmark buildings, distinctive structures, and location-specific features from map imagery to enable precise location identification within a city.

  \textbf{Task:} Analyze the provided map image and identify key landmark features that can help pinpoint the exact location within a city. Focus on:
  \begin{enumerate}
    \item Landmark buildings
    \item Major road names, intersections, and highway junctions
    \item Distinctive geographic features
    \item Notable commercial areas, industrial zones, or residential complexes with unique names
    \item Any visible place names or business names that serve as reference points
  \end{enumerate}

  \textbf{Output Format:} Exclude UI elements, map controls, and watermarks. Provide a 2-4 sentence description with specific landmark names and their relative positions to enable precise location identification. Use English only.

  \textbf{Example:} The location is at the intersection of Buena Vista Drive and South Drive, near Brothers Market. Calabasas Road lies to the northeast, with a large open space to the south.
\end{tcolorbox}

%% file: content/8_CameraRready.tex
% \clearpage
\appendix

\vspace{10pt}
\section{Further Analysis of JL-MMD}
\label{app:jlmmd-validation}

\subsection{Error Bound and Computational Efficiency}
JL-MMD evaluates generated dynamics after first encoding each network-traffic sample into a topology-aware representation. Given a graph $G=(V,\mathbf{A})$ and observations $\mathbf{X}$, the embedding function $\psi(G,\mathbf{X})$ aggregates temporal features through adjacency neighborhoods before graph-level pooling. Therefore, two samples with identical point-wise values but different connectivities can still differ in the JL-MMD space. The JL projection $\boldsymbol{\Phi}$ then reduces this representation while preserving pairwise distances over the sampled embedding set $\mathcal{S}$.

Let $\mathcal{S}$ contain $n$ sampled embeddings from $\mathcal{P}\cup\mathcal{Q}$, and let $\boldsymbol{\Phi}\in\mathbb{R}^{k\times d}$ be a JL projection with $k=O(\varepsilon^{-2}\log n)$. For any $L$-Lipschitz distance kernel $\kappa(\mathbf{x},\mathbf{y})=g(\|\mathbf{x}-\mathbf{y}\|_2^2)$, define $D_{\max}=\max_{\mathbf{u},\mathbf{v}\in\mathcal{S}}\|\mathbf{u}-\mathbf{v}\|_2$. With probability at least $1-2/n^2$,
\begin{equation}
\left|\mathrm{MMD}_{\kappa}^{2}(\boldsymbol{\Phi}\mathcal{P},\boldsymbol{\Phi}\mathcal{Q})-\mathrm{MMD}_{\kappa}^{2}(\mathcal{P},\mathcal{Q})\right|
\leq 4L\varepsilon D_{\max}^{2}+\delta_n,
\end{equation}
where the first term is projection distortion and $\delta_n=O(n^{-1/2})$ is finite-sample MMD estimation error.

The classical JL lemma states that all sampled pairs $\mathbf{u},\mathbf{v}\in\mathcal{S}$ preserve squared distance within $(1\pm\varepsilon)$:
\begin{equation}
(1-\varepsilon)\|\mathbf{u}-\mathbf{v}\|_2^2
\leq \|\boldsymbol{\Phi}\mathbf{u}-\boldsymbol{\Phi}\mathbf{v}\|_2^2
\leq (1+\varepsilon)\|\mathbf{u}-\mathbf{v}\|_2^2 .
\end{equation}
Thus the squared-distance perturbation is at most $\varepsilon D_{\max}^{2}$. Since $g$ is $L$-Lipschitz, each projected kernel entry changes by at most $B=L\varepsilon D_{\max}^{2}$:
\begin{equation}
\left|\kappa(\boldsymbol{\Phi}\mathbf{u},\boldsymbol{\Phi}\mathbf{v})-\kappa(\mathbf{u},\mathbf{v})\right|\leq B.
\end{equation}
The MMD expansion contains two within-distribution averages and one cross-distribution average,
\begin{equation}
\mathrm{MMD}^{2}_{\kappa}=\mathbb{E}_{\mathbf{x},\mathbf{x}'}\kappa(\mathbf{x},\mathbf{x}')+
\mathbb{E}_{\mathbf{y},\mathbf{y}'}\kappa(\mathbf{y},\mathbf{y}')-
2\mathbb{E}_{\mathbf{x},\mathbf{y}}\kappa(\mathbf{x},\mathbf{y}).
\end{equation}
Replacing every kernel entry by its projected counterpart changes these three terms by at most $B+B+2B=4B$, and the empirical-to-population gap contributes $\delta_n$. This explains why the final error has both a JL projection term and a finite-sample term.

For the linear kernel used by JL-MMD, the polarization identity rewrites $\langle\mathbf{x},\mathbf{y}\rangle=(\|\mathbf{x}\|_2^2+\|\mathbf{y}\|_2^2-\|\mathbf{x}-\mathbf{y}\|_2^2)/2$, so JL norm and distance preservation yield $L=1/2$ and $O(nd)$ evaluation. RBF satisfies the same form with $L=\gamma$, but requires $O(n^2d)$ pairwise evaluations.

\ifdefined\arxivjlmmdanalysisextra
\arxivjlmmdanalysisextra
\fi

\subsection{Synthetic Validation}
Here we further analyze JL-MMD using synthetic data to validate its sensitivity to topology-only changes. We generate 1000 random geometric graphs with 100 nodes and 24 time steps, and then compare the original with three types of controlled perturbations:
\begin{itemize}
    \item \textbf{Edge Rewiring (ER)} randomly rewires edges, changing local connectivity while keeping temporal dynamics unchanged. 
    \item \textbf{Community Rewiring (CR)} rewires edges across detected communities, producing a stronger meso-scale topology shift. 
    \item \textbf{Spatial-correlation Shuffling (SCS)} permutes the assignment between node signals and graph positions, preserving marginal values but breaking spatial correlations.
\end{itemize}

Empirical results in Table~\ref{tab:jlmmd_toy} show all existing metrics (i.e., ED, C-FID, MRR@5, MAE, RMSE, MAPE, WAPE) fail on topology-only perturbations, while JL-MMD can detect ER and CR with a high significance. JL-RBF behaves similarly, validating the topology sensitivity of the scalable linear-kernel JL-MMD used in the main experiments.

\input{tables/jlmmd_toy}

\ifdefined\arxivsyntheticvalidationextra
\arxivsyntheticvalidationextra
\fi

\input{figures/geograph}

\section{Qualitative Evaluation of Fidelity}
\label{app:geograph}

To further illustrate the topology-preserving capability of different models, we visualize the generated network states in three selected scenarios using heatmaps, as presented in Figure~\ref{fig:geograph}. Simulation results produced by Diffusion-TS and VerbalTS are less accurate and spatiallynoisy, failing to preserve the original network structures. ChatTraffic generates overly smoothed patterns that lose fine-grained spatial details. T2S produces more structured results but still exhibits artifacts in certain regions. In contrast, \method{} generates heatmaps that most closely resemble the ground truth, accurately capturing both the global distribution patterns and local variations across the networks. This visualization serves as empirical evidence that Graph U-Net architecture effectively preserves topological structure for network simulation.

%% file: tables/jlmmd_toy.tex
\begin{table}[H]
\centering
\caption{Synthetic validation of JL-MMD. ER and CR alter topology while preserving node values, and SCS shuffles spatial correlations. \textcolor{green!60!black}{$\checkmark$}: detected change ($>0.001$); \textcolor{red}{$\times$}: no detection.}
\label{tab:jlmmd_toy}
\resizebox{\columnwidth}{!}{
\begin{tabular}{l|cccc}
\toprule
\toprule
\textbf{Metric} & \textbf{Original} & \textbf{ER} & \textbf{CR} & \textbf{SCS} \\
\midrule
\textbf{ED} & 0.000 & 0.000 \textcolor{red}{$\times$} & 0.000 \textcolor{red}{$\times$} & 9.641 \textcolor{green!60!black}{$\checkmark$} \\
\textbf{C-FID} & 0.000 & 0.000 \textcolor{red}{$\times$} & 0.000 \textcolor{red}{$\times$} & 1.608$^{***}$ \textcolor{green!60!black}{$\checkmark$} \\
\textbf{MRR@5} & 1.000 & 1.000 \textcolor{red}{$\times$} & 1.000 \textcolor{red}{$\times$} & 1.000 \textcolor{red}{$\times$} \\
\textbf{MAE} & 0.000 & 0.000 \textcolor{red}{$\times$} & 0.000 \textcolor{red}{$\times$} & 0.317 \textcolor{green!60!black}{$\checkmark$} \\
\textbf{RMSE} & 0.000 & 0.000 \textcolor{red}{$\times$} & 0.000 \textcolor{red}{$\times$} & 0.448 \textcolor{green!60!black}{$\checkmark$} \\
\textbf{MAPE} & 0.000 & 0.000 \textcolor{red}{$\times$} & 0.000 \textcolor{red}{$\times$} & 277.419 \textcolor{green!60!black}{$\checkmark$} \\
\textbf{WAPE} & 0.000 & 0.000 \textcolor{red}{$\times$} & 0.000 \textcolor{red}{$\times$} & 0.269 \textcolor{green!60!black}{$\checkmark$} \\
\midrule
\textbf{JL-MMD} & 0.000 & 0.016$^{***}$ \textcolor{green!60!black}{$\checkmark$} & 0.081$^{***}$ \textcolor{green!60!black}{$\checkmark$} & 0.009$^{*}$ \textcolor{green!60!black}{$\checkmark$} \\
\textbf{JL-RBF} & 0.000 & 0.024$^{***}$ \textcolor{green!60!black}{$\checkmark$} & 0.116$^{***}$ \textcolor{green!60!black}{$\checkmark$} & 0.013$^{*}$ \textcolor{green!60!black}{$\checkmark$} \\
\bottomrule
\bottomrule
\end{tabular}
}
\end{table}

%% file: figures/geograph.tex
\begin{figure*}[!t]
    \centering
    \setlength{\tabcolsep}{2pt}
    \scriptsize
    \begin{tabular}{@{}cccccc@{}}
        Ground Truth & Diffusion-TS & VerbalTS & ChatTraffic & T2S & \method \\[2pt]
        \includegraphics[width=0.155\textwidth]{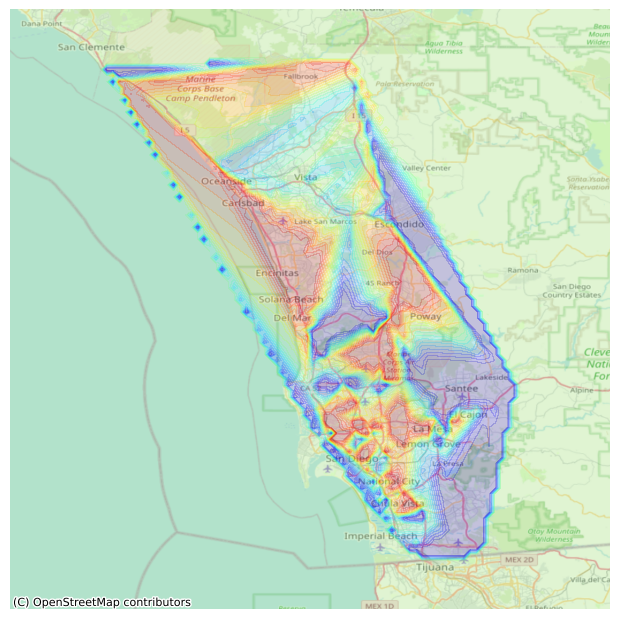} &
        \includegraphics[width=0.155\textwidth]{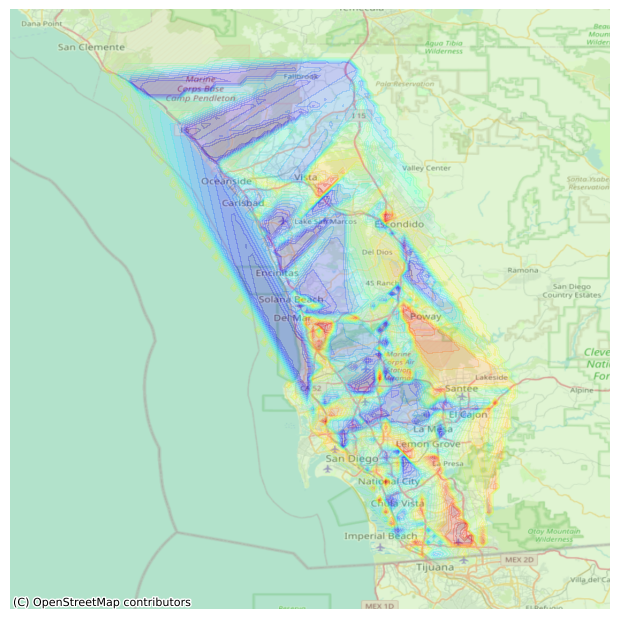} &
        \includegraphics[width=0.155\textwidth]{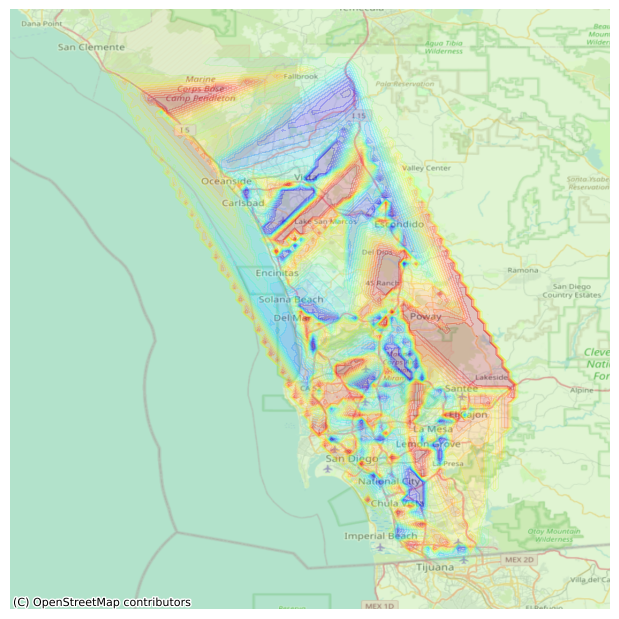} &
        \includegraphics[width=0.155\textwidth]{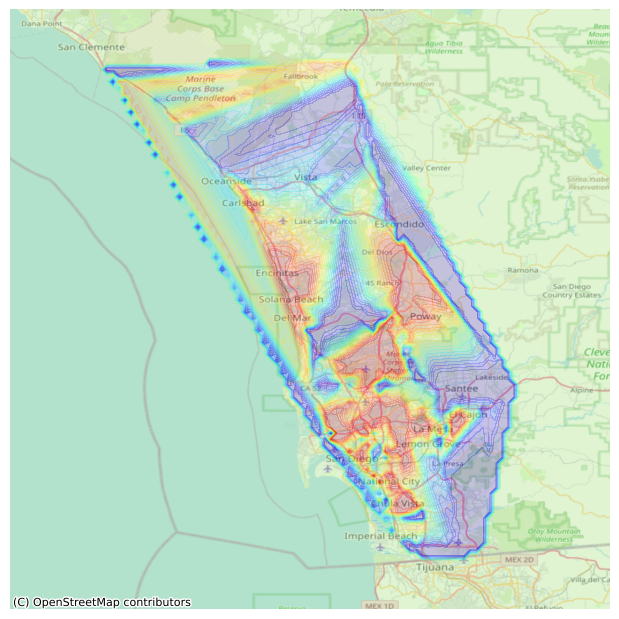} &
        \includegraphics[width=0.155\textwidth]{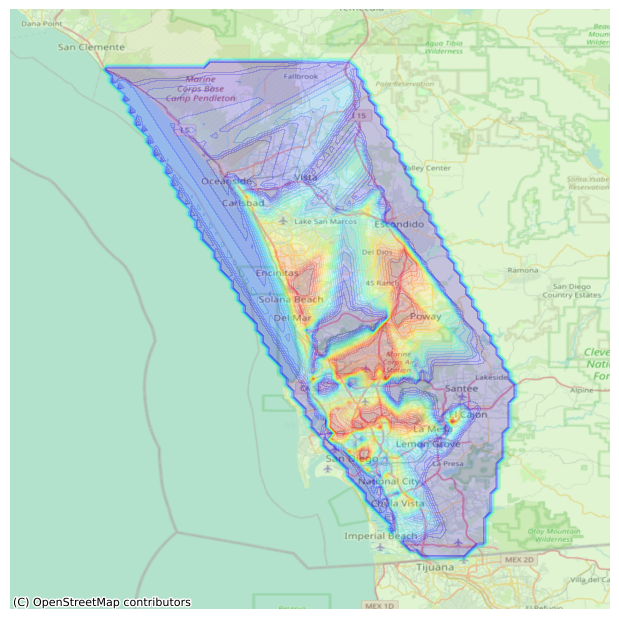} &
        \includegraphics[width=0.155\textwidth]{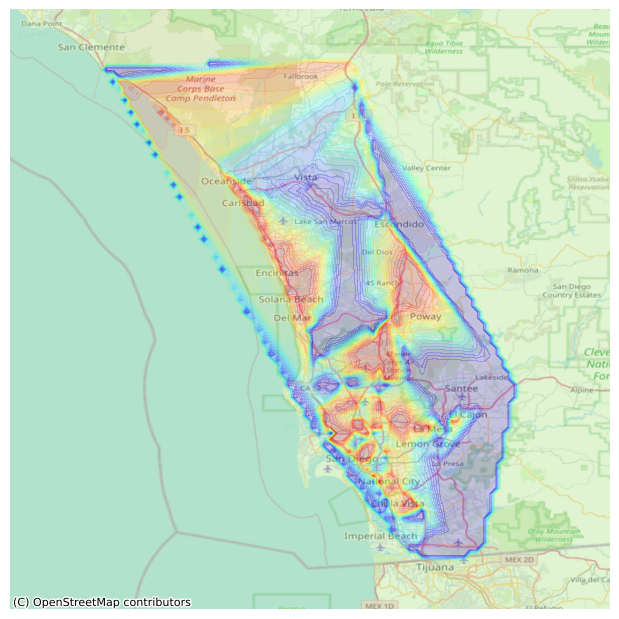} \\
        \multicolumn{6}{@{}p{0.98\textwidth}@{}}{\centering \textbf{(a) SD dataset (2017-01-05 20:10)}. Delays of three minutes on San Diego Fwy Northbound in San Diego.} \\[4pt]

        \includegraphics[width=0.155\textwidth]{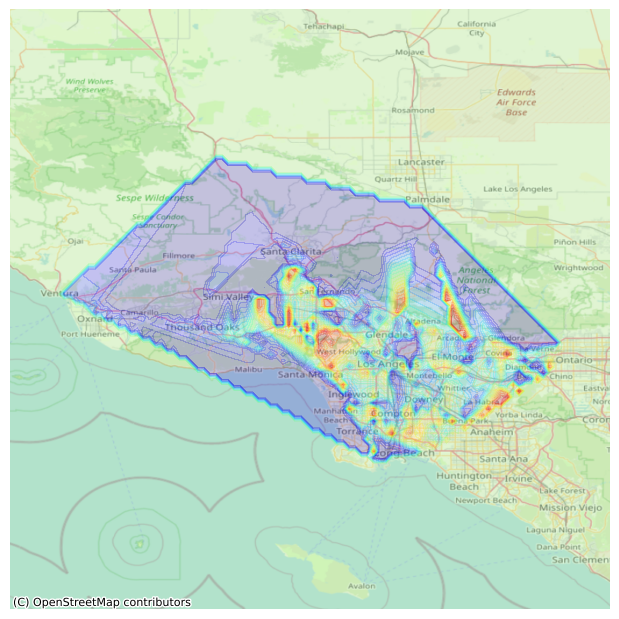} &
        \includegraphics[width=0.155\textwidth]{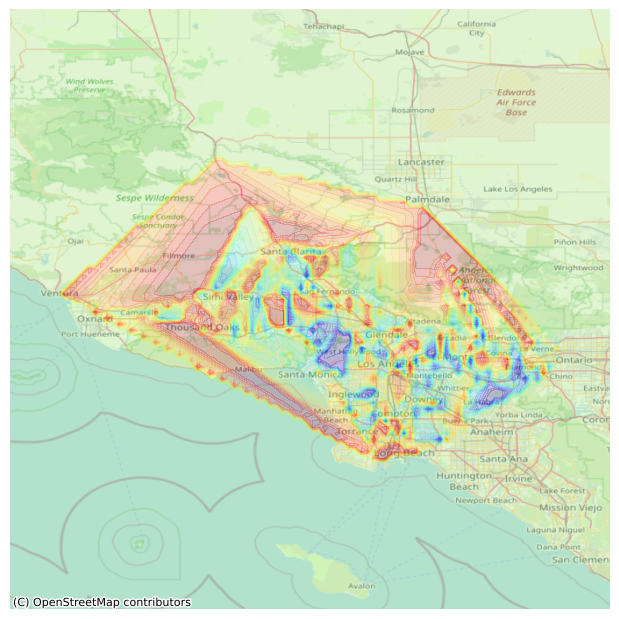} &
        \includegraphics[width=0.155\textwidth]{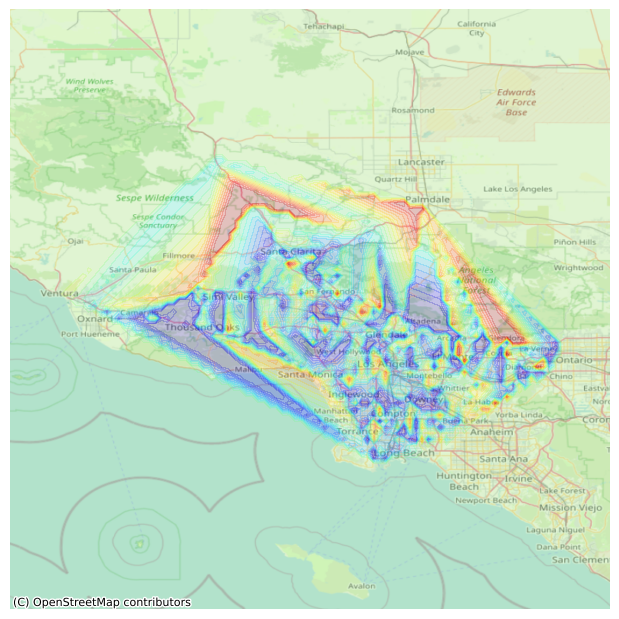} &
        \includegraphics[width=0.155\textwidth]{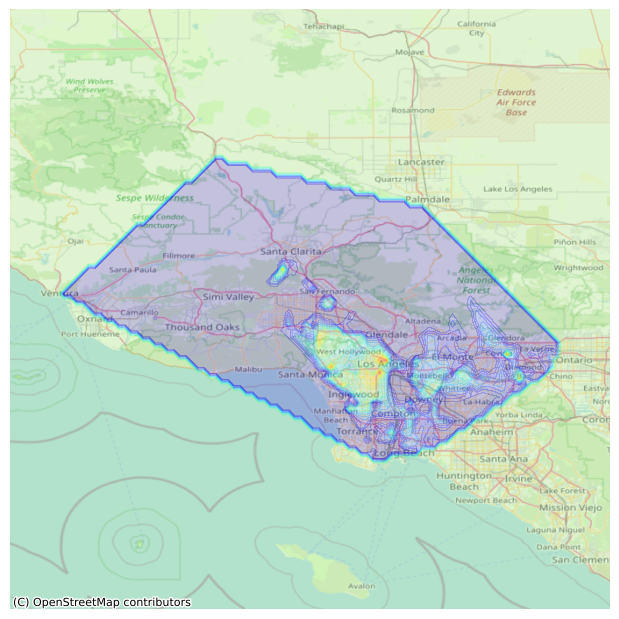} &
        \includegraphics[width=0.155\textwidth]{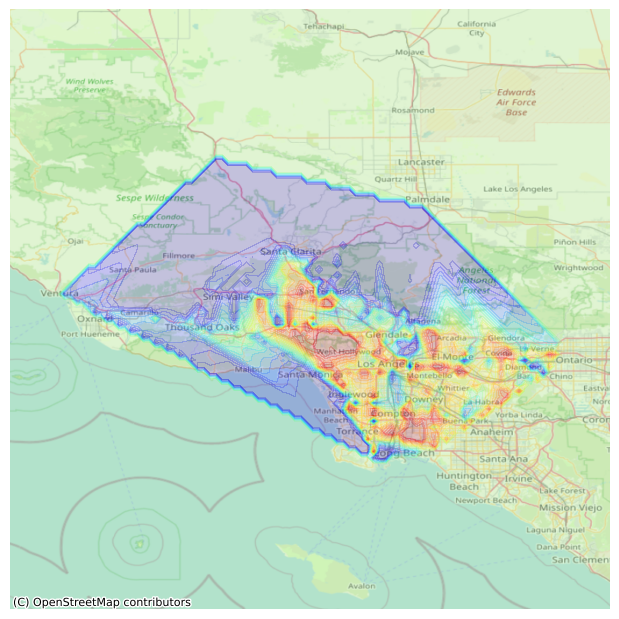} &
        \includegraphics[width=0.155\textwidth]{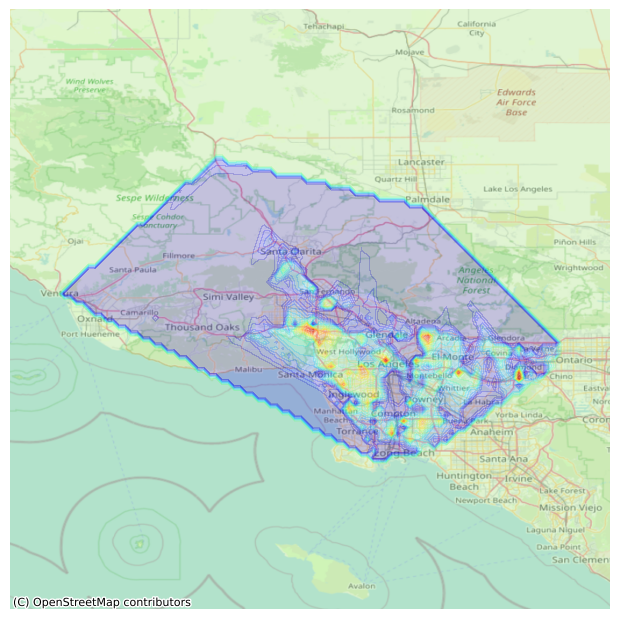} \\
        \multicolumn{6}{@{}p{0.98\textwidth}@{}}{\centering \textbf{(b) GLA dataset (2017-08-17 14:45)}. Delays of five minutes on Nimitz Fwy Southbound in Hayward. Average speed 20 mph. } \\[4pt]

        \includegraphics[width=0.155\textwidth]{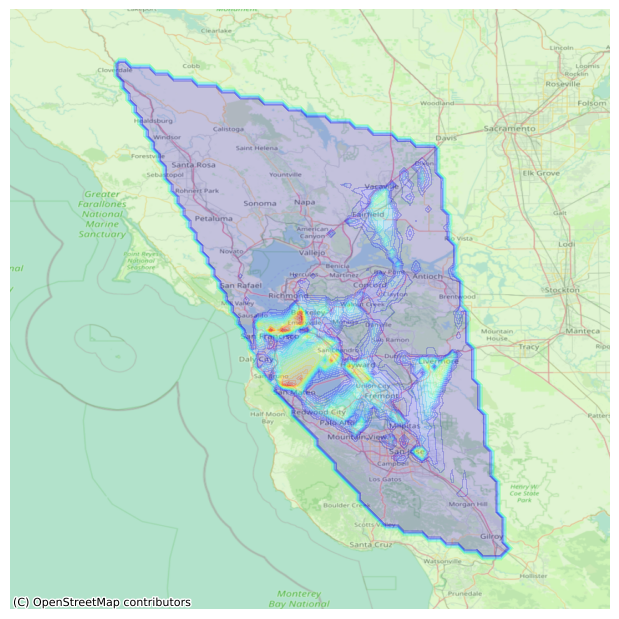} &
        \includegraphics[width=0.155\textwidth]{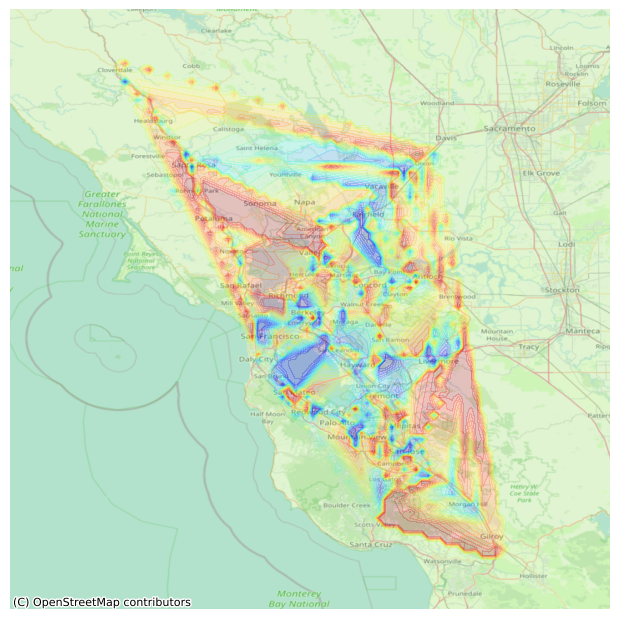} &
        \includegraphics[width=0.155\textwidth]{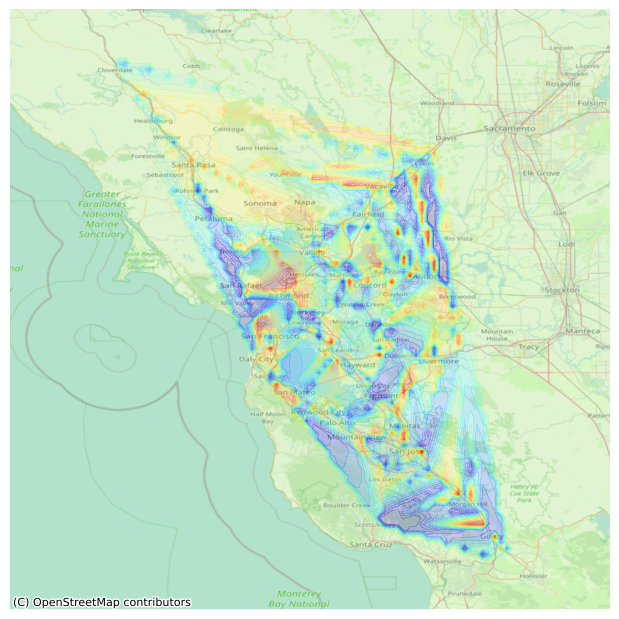} &
        \includegraphics[width=0.155\textwidth]{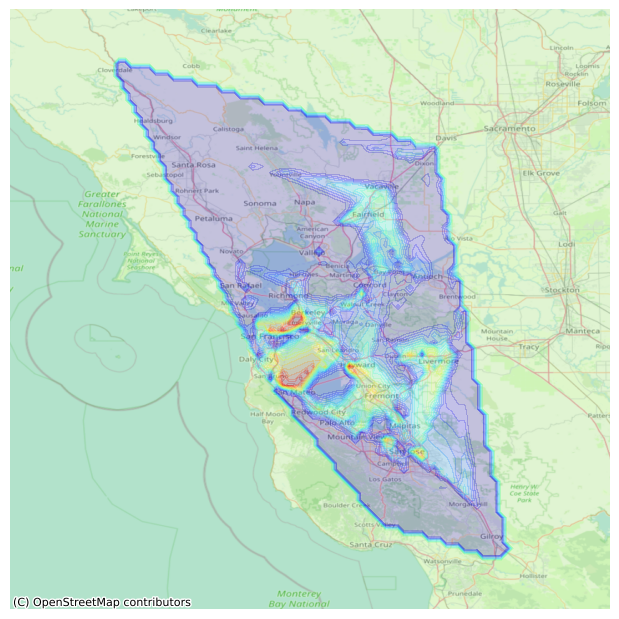} &
        \includegraphics[width=0.155\textwidth]{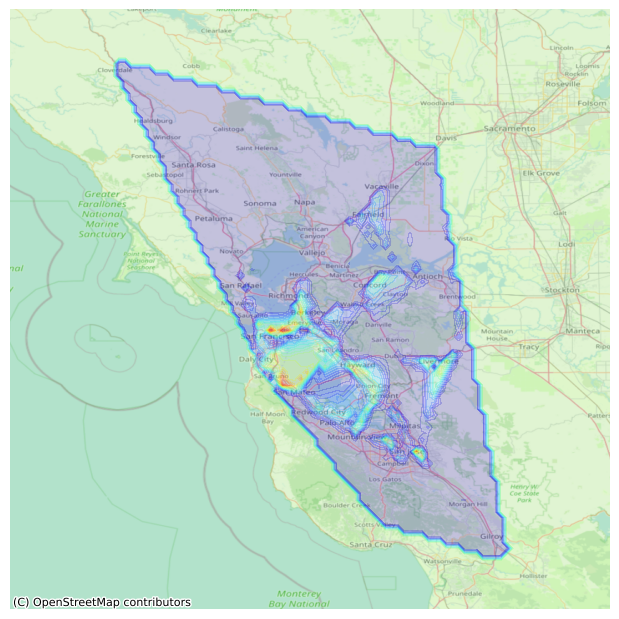} &
        \includegraphics[width=0.155\textwidth]{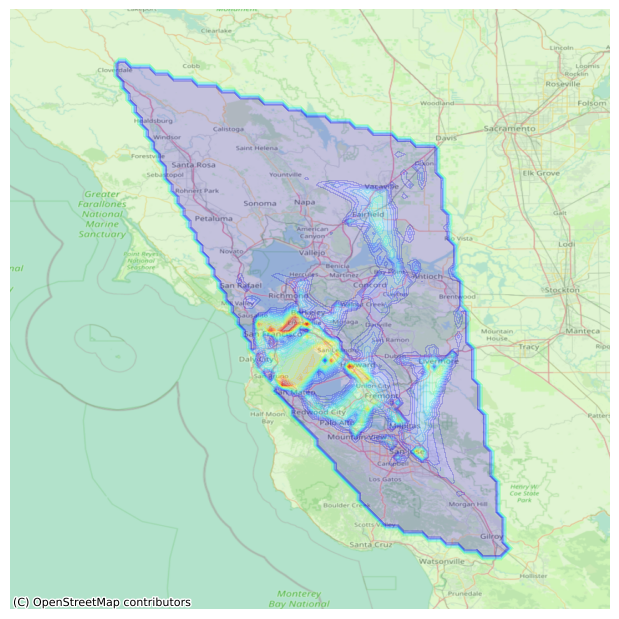} \\
        \multicolumn{6}{@{}p{0.98\textwidth}@{}}{\centering \textbf{(c) GBA dataset (2017-09-07 13:25)}. Delays of two minutes and delays easing on Artesia Fwy Eastbound between CA-19 Lakewood Blvd and CA-91. Average speed 25 mph.} \\
    \end{tabular}
    \caption{Qualitative comparison of generation results}
    \Description{Three rows compare real and generated geographical traffic heatmaps for SD, GLA, and GBA. Columns show the real map and predictions from Diffusion-TS, VerbalTS, ChatTraffic, T2S, and Net-Ev2, with Net-Ev2 visually closest to the road-aligned ground-truth patterns.}
    \label{fig:geograph}
\end{figure*}